%% file: main.tex
\documentclass{article}

\PassOptionsToPackage{numbers, compress}{natbib}

\usepackage[main, final]{neurips_2026}

\usepackage[utf8]{inputenc} 
\usepackage[T1]{fontenc}    
\usepackage{hyperref}       
\usepackage{url}            
\usepackage{booktabs}       
\usepackage{amsfonts}       
\usepackage{nicefrac}       
\usepackage{microtype}      
\usepackage{xcolor}         
\newcommand{\proposed}{CHASE}
\usepackage{enumitem}
\usepackage{graphicx}
\usepackage{wrapfig}
\usepackage{algorithm}
\usepackage{algorithmic}
\usepackage[table]{xcolor}
\usepackage{titletoc}
\hypersetup{linktocpage=true}
\newcommand\DoToC{%
  \startcontents
  \printcontents{}{1}{\vskip3pt\hrule\vskip5pt}
  \vskip3pt\hrule\vskip5pt
}

\title{Chasing Label Shifters: A Change-Aware Framework for Dynamic Graph Node Classification}

\author{
\textbf{Junghoon Kim}$^{1}$, 
\textbf{Seungyoon Choi}$^{1}$, 
\textbf{Hyunsung Kim}$^{1}$, 
\textbf{Chanyoung Park}$^{1}$\thanks{Corresponding Author} \\
$^{1}$KAIST \\
\texttt{\{jhkim611, csyoon08, hyunsung.kim, cy.park\}@kaist.ac.kr} \\
}

\begin{document}

\maketitle

\begin{abstract}
  Entities in real-world systems often evolve over time: users shift between information consumption patterns, researchers migrate between fields, and firms transition between financial risk states. When such systems are modeled as dynamic graphs, these transitions correspond to nodes whose class labels change between consecutive snapshots, which we call \textit{shifters}. Correctly classifying shifters is often more consequential than classifying stable nodes with persistent labels, as detecting such transitions enables timely intervention in high-stakes settings. Yet, we identify a systematic failure mode across all major dynamic graph neural network architectures: shifter performance consistently lags behind stable nodes, with errors concentrated on the old label. Our structural analysis traces this failure to \textit{embedding inertia}: neighborhood aggregation keeps shifter representations anchored to the old-class centroid, and this effect is strongest precisely where it is hardest to correct, namely for shifters whose neighborhoods remain aligned with the old class. Guided by this finding, we propose \textbf{\proposed}, a model-agnostic \textbf{\underline{CH}}ange-\textbf{\underline{A}}ware framework for \textbf{\underline{S}}hifting nod\textbf{\underline{E}}s that wraps any dynamic GNN with targeted components for detecting label shifts and overriding stale neighborhood signals. \proposed~consistently improves shifter performance across all tested models and datasets (up to 114.9\% and 202.5\% improvement on accuracy and F1 score, respectively) while preserving stable-node performance. We additionally contribute three dynamic graph benchmarks with naturally shifting labels, filling a gap in existing resources. The source code is available at \url{https://github.com/jhkim611/CHASE}.
\end{abstract}

\vspace{-3ex}
\section{Introduction}
\vspace{-1ex}
\begin{wrapfigure}{r}{0.6\textwidth}
  \centering
  \vspace{-20pt}
  \includegraphics[width=0.58\textwidth]{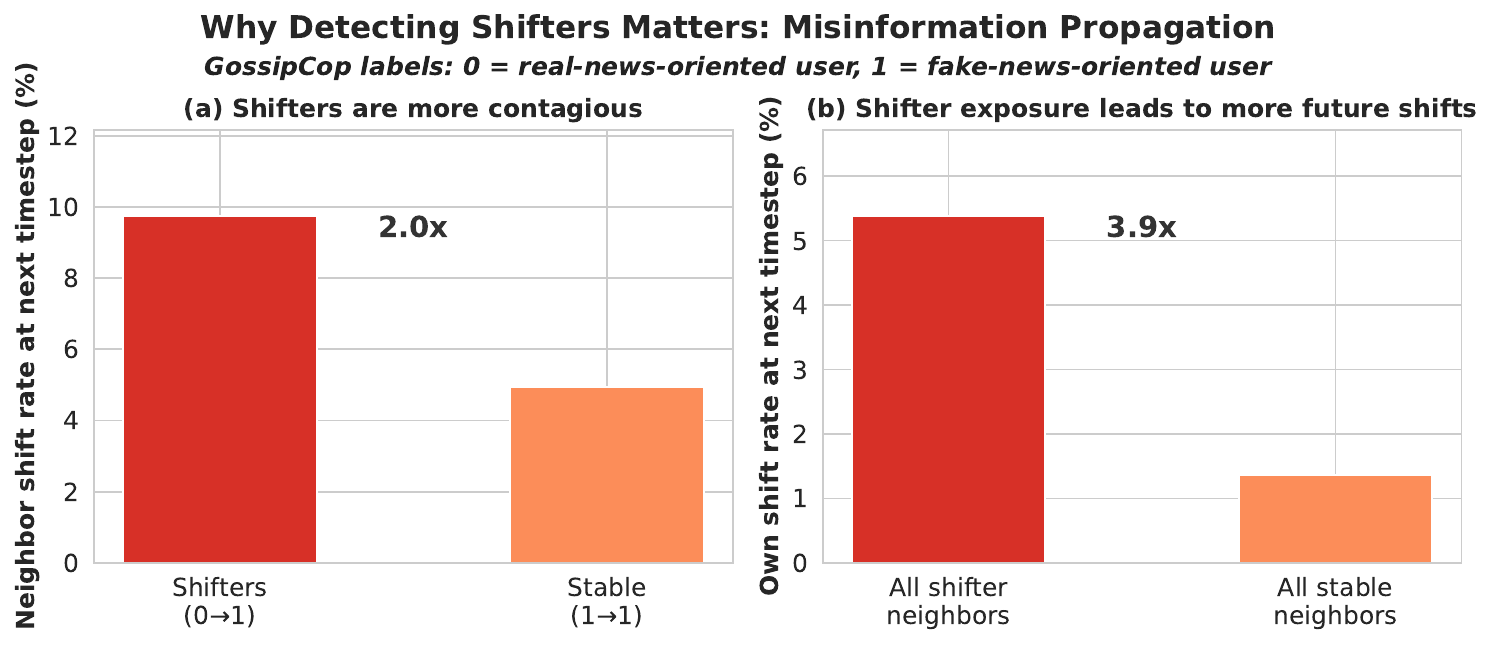}
  \vspace{-10pt}
  \caption{In misinformation propagation, recent \textit{shifters} are more influential than stable entities, making their misclassification much more critical.}
  \label{fig:motivation}
  \vspace{-10pt}
\end{wrapfigure}

In many real-world networks, the entities represented by nodes are not static~\cite{entitychange}. Researchers shift their primary area of investigation, social media users alter their information consumption and sharing patterns, listeners migrate across musical genres, and firms transition between financial risk states. These transitions arise both from intrinsic behavioral change and from evolving relational context: a researcher's new collaborations may pull them toward a different subfield, a user's changing social circle may expose them to different content, or a firm's shifting transaction network may signal deteriorating creditworthiness. We refer to nodes whose class label changes between consecutive time steps as \textbf{\textit{shifters}}, and to those whose labels persist as \textbf{\textit{stable nodes}}.

Such shifters are often disproportionately consequential, making their correct classification more critical than that of stable nodes. In misinformation detection, for instance, users tend to orient toward either credible or unreliable content~\cite{usernewsexposure, decor}, and the most operationally important nodes are not the stable consumers of fake news, but those who are actively transitioning toward fabricated articles. On the GossipCop dataset~\cite{fakenewsnet}, where users are connected by co-consumption of news articles and labeled according to whether the majority of their consumed content is fabricated, we find that real-news neighbors of a recent shifter toward fake news are 2.0$\times$ more likely to shift themselves than those of a stable fake-news user (See Fig.~1(a)). Moreover, this effect scales with the concentration of recent transitions in the neighborhood: among nodes with fake-news neighbors, those whose fake-news neighbors are predominantly recent shifters are $3.9\times$ more likely to shift than those whose fake-news neighbors are predominantly long-standing consumers (See Fig.~\ref{fig:motivation}(b)). In other words, recent shifters are \textit{more influential} than established members of the opposing class, with greater potential to cause future shifts in their neighbors. Hence, accurately detecting these transitions would enable timely intervention, e.g., identifying at-risk neighbors before they shift, a capability that correct classification of stable nodes alone cannot provide.

\textit{Dynamic graph neural networks}~\cite{dgsurvey1, dgsurvey2} are the natural modeling framework for such settings, as they jointly capture evolving topology and node features over time. These models typically combine a spatial graph encoder with a sequence model to transfer information across discrete snapshots, yet their architectures do not specifically target nodes whose labels change. Further, existing dynamic node classification benchmarks either lack node features~\cite{gcnse} or do not exhibit meaningful label change{~\cite{hydg,evolvegcn}}, making systematic evaluation of shifter performance difficult.

\looseness=-1
Therefore, to study label shift in this context, we construct three dynamic graph benchmarks from real-world data, from which we identify a systematic failure mode shared by all major dynamic GNN architectures. Our datasets span academic co-authorship (\textbf{DBLP-S}), news co-consumption (\textbf{GossipCop}), and music co-listening (\textbf{Music}), where node labels, derived from observed behavioral patterns, shift naturally over time. Across all three datasets and seven models, we find that \textbf{\textit{shifters are classified with substantially lower accuracy than stable nodes}} (See Fig.~\ref{fig:gap}(a)--(b)). The failure is directional: when models misclassify shifters, they overwhelmingly predict the old label (See Fig.~\ref{fig:gap}(c)). This pattern holds across all architectures examined and is not attributable to any single design choice.

To understand this failure, we conduct a structural analysis of shifter classification across all models (See Section~\ref{analysis}). We identify a single architecture-independent cause, which we term \textbf{\textit{embedding inertia}}: neighborhood aggregation keeps shifter representations anchored to the old-class centroid, and this anchoring is strongest for shifters whose local homophily (the fraction of neighbors sharing the same class) still reflects the old class, i.e., shifters with low current homophily and high previous homophily. Our analysis reveals that correctly classified shifters exhibit positive centroid displacement toward their new class, while incorrectly classified shifters show no such movement.

Guided by these findings, we propose \textbf{\proposed} (\textbf{\underline{CH}}ange-\textbf{\underline{A}}ware framework for \textbf{\underline{S}}hifting nod\textbf{\underline{E}}s), a model-agnostic augmentation that wraps any dynamic GNN model without modifying its encoder. \proposed~introduces three components, each addressing a specific diagnostic finding: a (a) \textbf{\textit{homophily-guided contrastive loss}} that pulls shifter embeddings toward their new-class centroid with strength inversely proportional to current homophily, counteracting embedding inertia where it is most severe; a (b) \textbf{\textit{change detection head}} that estimates the probability of a label shift from structural signals; and a (c) \textbf{\textit{change-aware classifier}} that receives the change detection estimate as an explicit signal for when to deviate from prior predictions, enabling the model to override stale neighborhood representations.

We evaluate \proposed~across seven models and three datasets. Shifter performance improves in all cases (up to 114.9\% and 202.5\% improvement on accuracy and F1 score, respectively), while stable-node performance is preserved or also enhanced. All components contribute monotonically in ablation. Together, these results establish label shift in dynamic graphs as a distinct, practically grounded, and largely underexplored problem, and demonstrate that it can be addressed through a model-agnostic framework without sacrificing performance on stable nodes.

Our contributions are as follows:
\begin{itemize}[leftmargin = 2mm]
    \item \textbf{The shifter gap on real-world dynamic graph benchmarks.} We construct three dynamic graph datasets from real-world data spanning academic collaboration, misinformation consumption, and music listening, in which node labels, derived from observed behavior, shift naturally across snapshots. Using these benchmarks, we identify and quantify a universal performance disparity on label-shifting nodes across seven dynamic GNN architectures, and attribute it to an architecture-independent mechanism (which we term \textit{embedding inertia}) in which neighborhood aggregation keeps shifter representations anchored to the old class, with the effect concentrated on shifters whose local homophily still reflects the old class.
    \item \textbf{Analysis-driven framework design.} Each component of \proposed~is motivated by a specific diagnostic finding, establishing a direct link between problem analysis and solution design. The contrastive loss corrects embedding inertia, the change detection head addresses \textit{when} to deviate, and the change-aware classifier provides the capacity for \textit{how} to deviate.
    \item \textbf{Model-agnostic augmentation with universal improvement.} \proposed~improves shifter classification across all backbone–dataset combinations while preserving stable-node performance. The framework requires no architectural modification to the base encoder, demonstrating that the shifter gap is addressable through targeted change awareness.
\end{itemize}

\section{Related works} \label{relatedworks}
\vspace{-1ex}
\subsection{Dynamic Graph Neural Networks}
\vspace{-1ex}
\textit{Dynamic graph neural networks} extend static GNN architectures to graphs whose topology, node features, or both evolve over time. The dominant paradigm for discrete-time dynamic graphs couples a spatial graph encoder with a sequence model. Early approaches integrate graph convolutions with recurrent units: GCRN~\cite{gcrn} applies GCN~\cite{gcn} followed by GRU~\cite{gru} to learn spatiotemporal representations, while EvolveGCN~\cite{evolvegcn} uses an RNN~\cite{gru,lstm} to evolve the GCN weight matrices themselves across snapshots. Attention-based alternatives replace the recurrent component with self-attention across snapshots: DySAT~\cite{dysat} applies structural GAT~\cite{gat} attention within each snapshot and transformer-style attention across snapshots, and GCN-SE~\cite{gcnse} learns squeeze-and-excitation attention weights over snapshot representations. More recent architectures explore implicit equilibrium formulations in IDGNN~\cite{idgnn}, time augmentation with deep propagation in TADGNN~\cite{tadgnn}, and biologically inspired spiking activation patterns in SiGNN~\cite{signn}.

While the dynamic graph literature broadly acknowledges that node properties, including labels, may evolve over time~\cite{evolvegcn,hydg,gcnse}, existing methods do not explicitly distinguish between \textit{shifters} (nodes whose labels change between consecutive snapshots) and \textit{stable nodes} (those whose labels persist), nor do they design components or evaluation protocols targeting label-shifting nodes. GCN-SE~\cite{gcnse} is the most explicit example: it acknowledges time-varying class membership as a property of dynamic graphs and observes on synthetic data that attention weights shift when labels change. However, its real-world evaluation on a dataset with changing labels uses a graph that lacks node features, and the model itself contains no mechanism that detects or adapts to label shifts. To the best of our knowledge, ours is the first work to isolate shifter performance as an evaluation criterion and introduce architectural components specifically designed for this purpose.

At the benchmark level, existing evaluation datasets for dynamic graph learning do not foreground label shift. The Temporal Graph Benchmark (TGB)~\cite{tgb} includes a node-level task, but it is node affinity prediction, i.e., ranking a node's future interactions, which is distinct from classifying nodes whose categorical labels change over time. The benchmark itself notes the scarcity of graph datasets with dynamic node labels. Our work addresses this gap by constructing datasets where node labels are derived from observed behavior and shift naturally across snapshots, and by explicitly evaluating model performance on shifters versus stable nodes.

\vspace{-1ex}
\subsection{Distinction from Distribution Shift on Graphs}
\vspace{-1ex}
A growing body of work addresses distribution shift in graph learning, where training and test distributions diverge due to differences in graph structure, feature distributions, or latent environments~\cite{disshift1, disshift2, disshift3, eerm, disshiftsurvey}. These methods assume a \textit{distributional} mismatch between training and deployment, and seek to learn invariant representations that generalize across contexts. On the other hand, the shifter problem is fundamentally different: it occurs \textit{within} the observed graph across consecutive time steps, where a known node's ground-truth label changes. The failure mode is not unfamiliar structure but \textit{temporal inertia}: the model's own representations and predictions resist updating when a node's class membership shifts. Addressing this requires detecting change and overriding learned continuity, rather than learning environment-invariant features.

\vspace{-1ex}
\section{Preliminaries and Datasets}
\vspace{-1ex}
\noindent{\textbf{Problem Setting.}}
We consider a discrete-time dynamic graph $\mathcal{G}=\{\mathcal{G}^{(1)},...,\mathcal{G}^{(T)}\}$, where each snapshot $\mathcal{G}^{(t)}=(V^{(t)},E^{(t)},X^{(t)},y^{(t)})$ consists of a node set $V^{(t)}$, edge set $E^{(t)}$, node feature matrix {$X^{(t)} \in \mathbb{R}^{|V^{(t)}| \times d}$} and label vector $y^{(t)}$. The node set may evolve across snapshots, where new nodes can appear and inactive nodes are retained but have zero degree~\cite{evolvegcn, idgnn, signn}. For a node $v$ present at consecutive snapshots $\mathcal{G}^{(t-1)}$ and $\mathcal{G}^{(t)}$, we define $v$ as a \textit{shifter} at snapshot $\mathcal{G}^{(t)}$ if $y_v^{(t)} \neq y_v^{(t-1)}$ and as \textit{stable} otherwise. We denote the set of shifters at snapshot $\mathcal{G}^{(t)}$ as $\mathcal{S}^{(t)}$ and stable nodes as $\bar{\mathcal{S}}^{(t)}$, with $n_{\mathrm{shifter}}^{(t)} = |\mathcal{S}^{(t)}|$ and $n_{\mathrm{stable}}^{(t)} = |\bar{\mathcal{S}}^{(t)}|$. The $T$ snapshots are divided into contiguous, non-overlapping training, validation, and test windows in temporal order. During training and validation, models have access to all available information (e.g., graph structure, node features, and labels) within the respective windows, but no test-time information is accessible at any point. At test time, the model predicts labels for all nodes in each test snapshot. Our analysis and method focus specifically on the disparity in classification performance between shifters and stable nodes.

\noindent{\textbf{Datasets.}}
We \textbf{\textit{construct three dynamic graph benchmarks from real-world data}}, spanning distinct domains and label semantics (See Table~\ref{tab:datasets}). In each case, node labels are derived from observed behavior and shift naturally across snapshots, i.e., they are not synthetically injected. Detailed construction procedures and feature descriptions are provided in Appendix~\ref{app:datasets}.

\begin{table}
\caption{Dataset statistics. Shifter \% is the fraction of nodes whose label differs from the previous snapshot, averaged across snapshots.}
\label{tab:datasets}
\centering
\resizebox{\columnwidth}{!}{
\begin{tabular}{lccccccc}
\toprule
Dataset & Domain & \# Nodes & \# Edges & Snapshots (Split) & Classes & Label meaning & Shifter \% \\
\midrule
DBLP-S & Academic & 34,717 & 394,385 & 14 (9/2/3) & 2 & Research area & 2.62\% \\
GossipCop & Social Media & 6,091 & 2,000,390 & 37 (26/5/6) & 2 & $>$50\% fake news consumption & 4.06\% \\
Music & Music Streaming & 878 & 1,009,457 & 24 (15/4/5) & 5 & Dominant genre & 18.77\% \\
\bottomrule
\end{tabular}
}
\vspace{-3ex}
\end{table}

\vspace{-2ex}
\begin{itemize}[leftmargin = 2mm]
    \item \textbf{DBLP-S.} (Source: DBLP-v11\footnote{\url{https://dblp.org/}})
Nodes are researchers and edges represent co-authorship within each yearly snapshot. Each researcher is labeled by their primary research area in that year. Label shifts correspond to researchers whose main field of study changes between consecutive years.
    \item \textbf{GossipCop.} (Source: FakeNewsNet~\cite{fakenewsnet})
Nodes are social media users and edges connect users who co-consume the same news article within a weekly snapshot. A user is labeled as a fake-news consumer if more than 50\% of the articles they engage with in that week are fabricated. Label shifts correspond to users whose consumption transitions from predominantly real to predominantly fake news, or vice versa.
    \item \textbf{Music.} (Source: LastFM-song-listens~\cite{lastfm1, lastfm2} and the Million Song Dataset~\cite{millionsong})
Nodes are listeners and edges connect users who listen to the same track within a monthly snapshot. Each listener is labeled by their dominant genre (out of five), determined by their listening history in that month. Label shifts correspond to listeners whose primary genre changes between consecutive months.
\end{itemize}

\vspace{-2ex}
\section{Diagnosing the Shifter Gap} \label{analysis}
\vspace{-1ex}
We evaluate seven dynamic GNN models on all three datasets: GCN-GRU~\cite{gcrn}, EvolveGCN~\cite{evolvegcn}, DySAT~\cite{dysat}, GCN-SE~\cite{gcnse}, IDGNN~\cite{idgnn}, TADGNN~\cite{tadgnn}, and SiGNN~\cite{signn}. For each node $v$ at snapshot $\mathcal{G}^{(t)}$, we compute the following structural signals. \textbf{\textit{Feature drift}}: $\mathrm{self\_drift}_v^{(t)}=\|\mathbf{x}_v^{(t)}-\mathbf{x}_v^{(t-1)}\|$, the change in $v$'s own features. \textbf{\textit{Neighborhood drift}}: $\mathrm{nbr\_drift}_v^{(t)}=\|\bar{\mathbf{x}}_{\mathcal{N}_v^{(t)}}^{(t)} - \bar{\mathbf{x}}_{\mathcal{N}_v^{(t)}}^{(t{-}1)}\|$, the change in the mean feature vector of $v$'s neighbors ($\mathcal{N}_v^{(t)}$ is $v$'s one-hop neighborhood at time $t$). \textbf{\textit{Current homophily}}: ${h_\mathrm{curr}}(v,t) = |\{u \in \mathcal{N}_v^{(t)} : y_u^{(t)} = y_v^{(t)}\}| / |\mathcal{N}_v^{(t)}|$, the fraction of $v$'s current neighbors sharing its current label. \textbf{\textit{Previous homophily}}: $h_{\mathrm{prev}}(v,t)={h_\mathrm{curr}}(v,t-1)$, the fraction of $v$'s previous neighbors sharing its old label. Meanwhile, to measure whether a node's learned representation has moved toward the correct class, we define \textbf{\textit{centroid displacement}}: $\Delta_c(v,t) = \mathrm{sim}(\mathbf{h}_v^{(t)}, \mathbf{c}_{v,\mathrm{new}}^{(t)}) - \mathrm{sim}(\mathbf{h}_v^{(t)}, \mathbf{c}_{v,\mathrm{old}}^{(t)})$, where $\mathrm{sim}$ denotes cosine similarity, $\mathbf{h}_v^{(t)}$ is the learned embedding of node $v$ at time $t$, and $\mathbf{c}_{v,\mathrm{new}}^{(t)}$ and $\mathbf{c}_{v,\mathrm{old}}^{(t)}$ are the class centroids of the new (at time $t$) and old (at time $t-1$) labels of $v$, respectively, computed from stable-node embeddings.

\begin{figure}[t]
\centering
\includegraphics[width=0.9\linewidth]{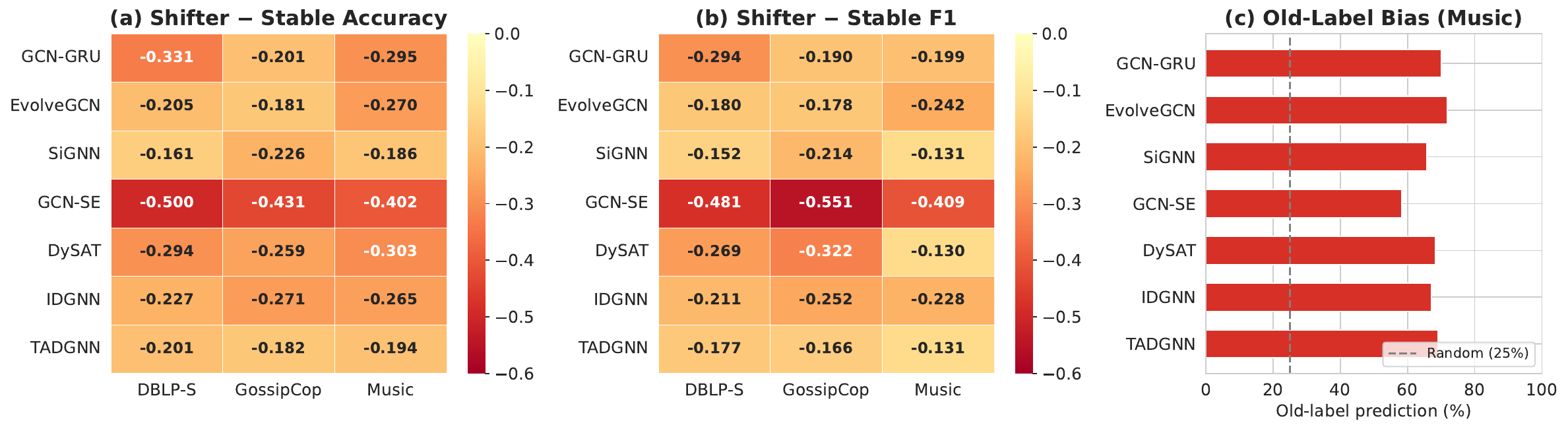}
\vspace{-2.5ex}
\caption{(a), (b) The shifter performance gap is universal across all models and datasets. (c) Shifter errors are predominantly stuck in the previous snapshot's label.}
\vspace{-3ex}
\label{fig:gap}
\end{figure}

\begin{figure}[t]
\centering
\includegraphics[width=0.9\linewidth]{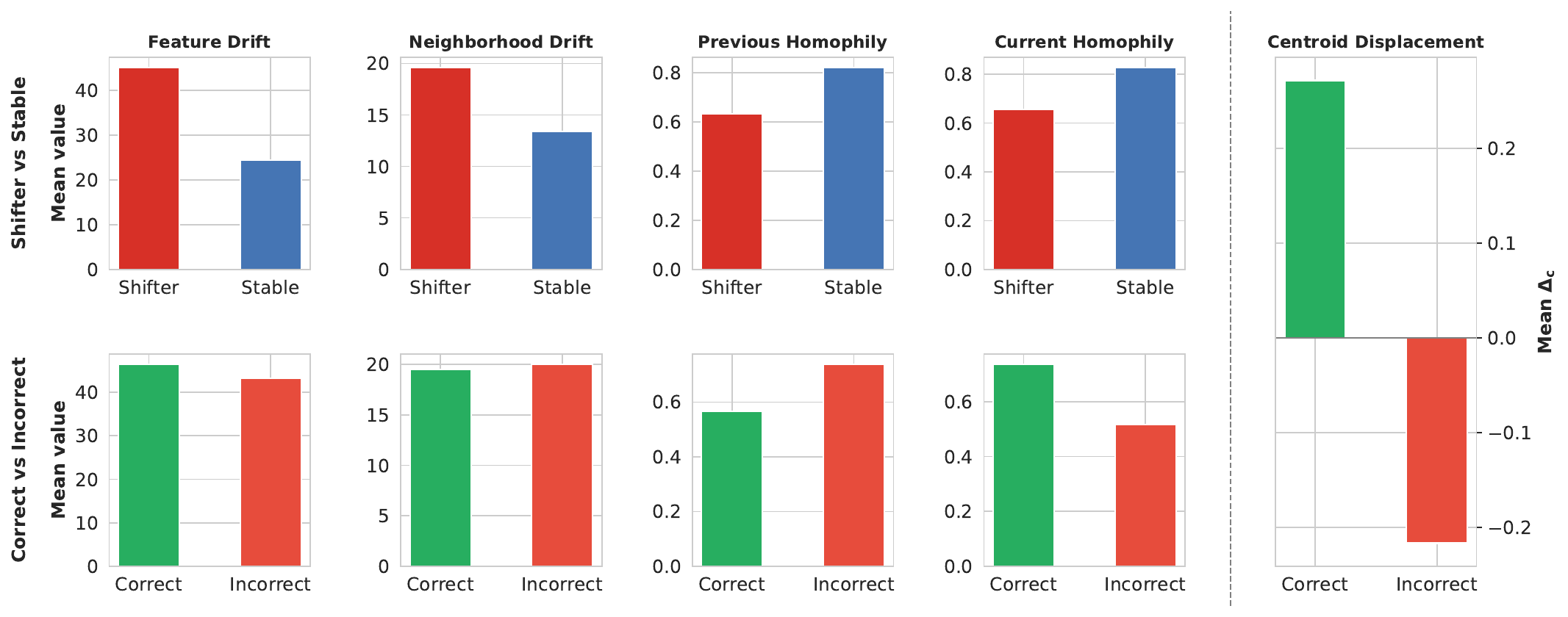}
\vspace{-2ex}
\caption{(Left) All four signals distinguish shifters from stable nodes (Top row), but only homophily signals meaningfully separate correctly from incorrectly classified shifters (Bottom row) (Mann-Whitney U test, $p \approx 0$ in all cases). (Right) Correctly classified shifters show positive centroid displacement while incorrectly classified ones do not, confirming embedding inertia. Averaged across models and datasets; per-model and per-dataset breakdowns in Appendix~\ref{app:analysis}.}
\vspace{-4ex}
\label{fig:signals}
\end{figure}

\noindent{\textbf{The shifter gap.}}
In Fig.~\ref{fig:gap}, we observe that shifters are classified with substantially lower performance than stable nodes across all seven models and three datasets. More precisely, shifter accuracy and F1 lag behind their stable-node counterparts by up to 50\%p and 55.1\%p, respectively (See Fig.~\ref{fig:gap}(a)--(b)). Moreover, these errors are not random: when models misclassify shifters, they overwhelmingly predict the \textit{old} label. On the multiclass \textbf{Music} dataset (5 classes), up to 76\% of shifter errors default to the old class, far exceeding the 25\% rate expected under uniform error (See Fig.~\ref{fig:gap}(c)). This old-label bias is consistent across all architectures and is not attributable to any single design choice.

\looseness=-1
\noindent{\textbf{Structural signals distinguish shifters from stable nodes.}}
To understand the sources of this failure, we first investigate which observable signals separate shifters from stable nodes (See Fig.~\ref{fig:signals}(Left), top row). We observe that feature drift and neighborhood drift are both significantly elevated for shifters. Shifters also exhibit lower current homophily and previous homophily than stable nodes. All four signals reliably identify \textit{which} nodes are shifters (rank-biserial $|r|$ from Mann-Whitney U test up to 0.94).

\noindent{\textbf{Within shifters, homophily determines correctness.}}
While these signals reliably identify shifters, a separate question is what determines whether a shifter is classified \textit{correctly} (See Fig.~\ref{fig:signals}(Left), bottom row). Here, the drift signals lose their discriminative power: feature drift and neighborhood drift do not meaningfully distinguish correctly from incorrectly classified shifters within the shifter subpopulation ($|r|$< 0.10). On the other hand, it turns out that the signals that \textit{do} distinguish them are homophily measures. Across all models and datasets, incorrectly classified shifters have significantly lower current homophily, i.e., their neighborhoods have not yet adopted the new label, and significantly higher previous homophily, i.e., they were deeply embedded in the old class ($|r|$ up to 0.52).

\noindent{\textbf{Embedding inertia.}}
These findings converge on a single failure mechanism. When a node shifts, neighborhood aggregation, the core operation of all GNN architectures, continues to produce representations anchored to the old class, because the local neighborhood still predominantly reflects the old label. This effect is strongest precisely where the homophily pattern is most unfavorable: low current homophily means the neighborhood provides no support for the new class, while high previous homophily means the old-class signal was strong before the shift. We term this mechanism \textit{embedding inertia}. The centroid displacement analysis confirms this directly (See Fig.~\ref{fig:signals}(Right)): correctly classified shifters exhibit positive $\Delta_c$, indicating their embeddings have moved toward the new-class centroid, while incorrectly classified shifters show $\Delta_c \approx 0$ or negative, meaning their embeddings have not moved at all ($p \approx 0$ across all models and datasets). Taking a closer look regarding the old-label bias, on \textbf{Music}, each shifter's embedding is most similar to its predicted class centroid (mean cosine similarity 0.83 vs.\ 0.65 to other classes). Shifters misclassified as their old class lie closest to that old class (0.84 vs.\ 0.78 to the true new class and 0.70 to the remaining three), whereas those misclassified as another class lie closest to that class (0.84 vs.\ 0.76 to the old class). Old-label errors, the majority of shifter errors, thus stem from embedding inertia toward the specific old class rather than from generic confusion.

\looseness=-1
This analysis yields three actionable insights for method design. First, embedding inertia is the core failure: shifter representations must be actively pulled toward the new class, with stronger correction where homophily is least supportive. Second, our observed structural signals effectively detect shifts, and should inform a dedicated detection mechanism whose output signals the rest of the model. Third, the old-label bias means that models default to the old class even when internal signals indicate change. As such, the classifier must receive an explicit change signal and the capacity to override prior predictions. These insights directly motivate the design of \proposed, described in the following section.

\vspace{-1ex}
\section{\proposed: \underline{CH}ange-\underline{A}ware Framework for \underline{S}hifting Nod\underline{E}s}
\vspace{-1ex}
\begin{wrapfigure}{r}{0.65\textwidth}
  \centering
  \vspace{-18pt}
  \includegraphics[width=0.55\textwidth]{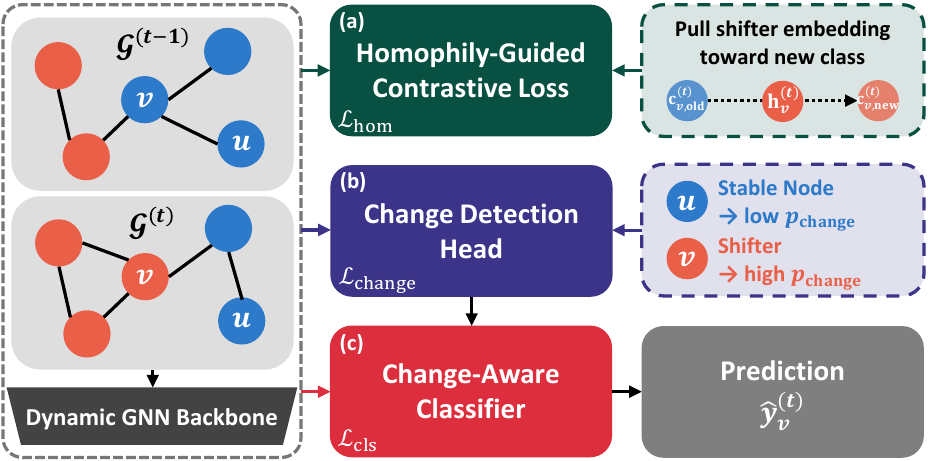}
  \vspace{-6pt}
  \caption{The \proposed~architecture. Given snapshots $\mathcal{G}^{(t-1)},\mathcal{G}^{(t)}$ and a dynamic GNN backbone, the \textbf{\textcolor[RGB]{8,80,65}{(a) homophily-guided contrastive loss}} pulls shifter embeddings toward their new-class centroid. Then, the \textbf{\textcolor[RGB]{60,52,137}{(b) change detection head}} estimates the probability $p_{\mathrm{change}}$ that a node has shifted. Finally, the \textbf{\textcolor[RGB]{220,46,61}{(c) change-aware classifier}} receives change signals to guide its prediction.}
  \label{fig:architecture}
  \vspace{-35pt}
\end{wrapfigure}

\proposed~is a model-agnostic augmentation that can wrap any dynamic GNN backbone without modifying its encoder. It replaces only the classifier head, leaving the base encoder's architecture untouched. At each snapshot $\mathcal{G}^{(t)}$, the backbone model produces embeddings $\mathbf{h}_v^{(t)}$ for each node as usual. Our proposed \proposed~augments the classification pipeline with three targeted components (See Fig.~\ref{fig:architecture}), each motivated by a specific finding from Section~\ref{analysis}.

\subsection{Homophily-Guided Contrastive Loss}
\vspace{-1ex}
\looseness=-1
Embedding inertia keeps shifter representations anchored to the old-class centroid (See Section~\ref{analysis}). To counteract this, we introduce a contrastive loss that pulls shifter embeddings toward their new-class centroid and pulls them away from the old. For each shifter $v$ at training time, we compute class centroids $\mathbf{c}_{v,\mathrm{new}}^{(t)}$ and $\mathbf{c}_{v,\mathrm{old}}^{(t)}$ from stable-node embeddings, which serve as reliable references since their labels have not changed. The loss is margin-based and averaged over all shifters across all training snapshots. Let $\mathcal{S} = \{(v, t) : v \in \mathcal{S}^{(t)},\, t \le \mathcal{T}_{\mathrm{train}}\}$ denote the set of all shifters across training. Then:
\begin{equation}
\mathcal{L}_{\mathrm{hom}} = \frac{1}{|\mathcal{S}|} \sum_{(v,t) \in \mathcal{S}} (1 - h_{\mathrm{curr}}(v, t)) \cdot \max\!\big(0,\; m - (\mathrm{sim}(\mathbf{h}_v^{(t)}, \mathbf{c}_{v,\mathrm{new}}^{(t)}) - \mathrm{sim}(\mathbf{h}_v^{(t)}, \mathbf{c}_{v,\mathrm{old}}^{(t)}))\big),
\end{equation}
where $\mathrm{sim}$ denotes cosine similarity, and $m$ is a margin hyperparameter (set to 0.5 across all experiments). The weighting by $(1 - h_{\mathrm{curr}}(v, t))$ applies stronger correction to shifters with low current homophily, precisely those for whom embedding inertia is most severe. This loss is applied only during training, where current labels are available.

\subsection{Change Detection Head}
\vspace{-1ex}
All four structural signals ($\mathrm{self\_drift}_v^{(t)}$, $\mathrm{nbr\_drift}_v^{(t)}$, $h_{\mathrm{curr}}(v, t)$ and $h_{\mathrm{prev}}(v, t)$) reliably detect shifters (See Section~\ref{analysis}), but this detection is not directly available to the classifier. The change detection head estimates the probability that a node has shifted:
\begin{equation}
p_{\mathrm{change}}(v, t) = \sigma\!\big(\mathrm{MLP}([\mathbf{h}_v^{(t)};\; \mathrm{self\_drift}_v^{(t)};\; \mathrm{nbr\_drift}_v^{(t)};\; h_{\mathrm{proxy}}(v, t)])\big),
\end{equation}
where our signals are concatenated. The term $h_{\mathrm{proxy}}(v, t)$ is a prediction-based approximation of previous homophily: $h_{\mathrm{proxy}}(v, t) = \mathrm{mean}_{u \in \mathcal{N}_v^{(t{-}1)}} \hat{\mathbf{y}}_u^{(t{-}1)}[\arg\max \hat{\mathbf{y}}_v^{(t{-}1)}]$, i.e., the average predicted probability that $v$'s previous neighbors share $v$'s predicted class at $t{-}1$. As true labels from the previous snapshot are not available during inference, we use this proxy consistently across training, validation, and test time, ensuring no label leakage. Note that current homophily $h_{\mathrm{curr}}$ is not used in the detection head: it requires current labels, which are the prediction target, and its role is already served by the contrastive loss weighting. {The head is supervised with weighted binary cross-entropy: $\mathcal{L}_{\mathrm{change}} = -\frac{1}{n}\sum_{v,t} \big[w^+ s_v^{(t)} \log p_{\mathrm{change}}(v, t) + (1 - s_v^{(t)}) \log (1 - p_{\mathrm{change}}(v, t))\big]$, where $s_v^{(t)} = 1[y_v^{(t)} \neq y_v^{(t-1)}]$ and $w^+ = n_{\mathrm{stable}} / n_{\mathrm{shifter}}$.} The output $p_{\mathrm{change}}$ is then passed to the classifier defined below as a soft shift signal. The head itself demonstrates considerable performance in detecting shifters, discussed in Appendix~\ref{app:further}.1.

\subsection{Change-Aware Classifier}
\vspace{-1ex}
The analysis shows that when models misclassify shifters, they default to the old label even when internal representations contain signals of change. The change-aware classifier is designed to receive these signals explicitly and learn when to override prior predictions. It takes as input:
\begin{equation}
\hat{\mathbf{y}}_v^{(t)} = \mathrm{MLP}\!\big([\mathbf{h}_v^{(t)};\; \mathbf{z}_{\mathrm{self},v}^{(t)};\; p_{\mathrm{change}}(v, t);\; \hat{\mathbf{y}}_v^{(t{-}1)}]\big),
\end{equation}
where $\mathbf{h}_v^{(t)}$ is the base encoder embedding (neighbor-aggregated), $\mathbf{z}_{\mathrm{self},v}^{(t)} = \mathrm{ReLU}(\mathbf{W}_{\mathrm{self}} \mathbf{x}_v^{(t)})$ is an aggregation-free projection of the node's own features, $p_{\mathrm{change}}(v, t)$ is the change detection output, and $\hat{\mathbf{y}}_v^{(t{-}1)}$ is the model's prediction from the previous snapshot. Each input serves a distinct role. When $p_{\mathrm{change}}$ is low (stable node), all signals tend to agree and the classifier can rely on $\mathbf{h}_v^{(t)}$ and $\hat{\mathbf{y}}_v^{(t{-}1)}$. When $p_{\mathrm{change}}$ is high (likely shifter), $\mathbf{h}_v^{(t)}$ and $\hat{\mathbf{y}}_v^{(t{-}1)}$ may still point to the old class, but $\mathbf{z}_{\mathrm{self},v}^{(t)}$ provides aggregation-free evidence and $p_{\mathrm{change}}$ signals the need to deviate. The MLP learns this interaction end-to-end, supervised with cross-entropy: $\mathcal{L}_{\mathrm{cls}} = \frac{1}{n}\sum_{v,t} w_{v}^{(t)} \,\ell_{\mathrm{CE}}(\hat{\mathbf{y}}_v^{(t)}, y_v^{(t)})$, where $w_{v}^{(t)} = \sqrt{(n_{\mathrm{shifter}}^{(t)} + n_{\mathrm{stable}}^{(t)}) / (2 \cdot n_{g(v,t)})}$ with $n_{g(v,t)} \in \{n_{\mathrm{shifter}}^{(t)}, n_{\mathrm{stable}}^{(t)}\}$ applies sqrt-dampened weights to address the rarity of shifters.

\subsection{Training Objective}
\vspace{-1ex}
The total loss combines the three terms: $\mathcal{L} = \mathcal{L}_{\mathrm{cls}} + \lambda_{\mathrm{hom}} \cdot \mathcal{L}_{\mathrm{hom}} + \lambda_{\mathrm{change}} \cdot \mathcal{L}_{\mathrm{change}}$, where $\lambda_{\mathrm{hom}} = \lambda_{\mathrm{change}} = 0.1$ across all experiments. \proposed~is trained end-to-end with the backbone in a single run, with all parameters, including the base encoder, updated jointly under $\mathcal{L}$. During inference, only the change detection head and the change-aware classifier are used, and $\mathcal{L}_{\mathrm{hom}}$ does not apply. All inputs at test time are derived from node features, graph structure, and the model's own previous predictions, \textit{requiring no ground-truth labels from any test snapshot}.

\subsection{Model-Agnostic Design}
\vspace{-1ex}
\looseness=-1
\proposed~replaces only the classifier head of the backbone model, i.e., no architectural modification to the base encoder. The baseline configuration maps $\mathbf{h}_v^{(t)} \to \mathrm{MLP} \to \hat{\mathbf{y}}_v^{(t)}$; \proposed~augments the input to $[\mathbf{h}_v^{(t)}; \mathbf{z}_{\mathrm{self},v}^{(t)}; p_{\mathrm{change}}(v, t); \hat{\mathbf{y}}_v^{(t{-}1)}] \to \mathrm{MLP} \to \hat{\mathbf{y}}_v^{(t)}$. Here, ``model-agnostic'' refers to the freedom to choose the backbone, not wrapping a frozen pre-trained model. All backbones use the same 2-layer MLP classifier, ensuring that performance differences are attributable to the \proposed~components rather than classifier capacity. Full pseudocode and a complexity analysis are provided in Appendix~\ref{app:method}.

\vspace{-1ex}
\section{Experiments}
\vspace{-1ex}
\subsection{Setup}
\vspace{-1ex}
\looseness=-1
We construct three dynamic graph datasets containing label shifts from real-world data spanning academic collaboration (\textbf{DBLP-S}), misinformation consumption (\textbf{GossipCop}), and music listening (\textbf{Music}) (See Table~\ref{tab:datasets} and Appendix~\ref{app:datasets}) for evaluating \proposed. We compare against seven dynamic GNN baselines: GCN-GRU~\cite{gcrn}, EvolveGCN~\cite{evolvegcn}, DySAT~\cite{dysat}, GCN-SE~\cite{gcnse}, IDGNN~\cite{idgnn}, TADGNN~\cite{tadgnn} and SiGNN~\cite{signn}. We additionally include two static baselines, i.e., a standard GCN~\cite{gcn} and EERM~\cite{eerm} (a GNN method designed to address distribution shift), in two settings: \textit{collapsed}, where all training snapshots are merged into a single graph, and \textit{sequential}, where the model is trained on each snapshot in order. The sequential variant is additionally augmented with \proposed. For each backbone, we compare the baseline classifier against the same encoder augmented with \proposed. All models are trained with five random seeds and selected based on validation F1. We report accuracy and macro F1, computed separately for shifters and stable nodes, averaged across test snapshots. Full implementation details, hyperparameters, and hardware specifications are provided in Appendix~\ref{app:implementation}.

\subsection{Main Results}
\vspace{-1ex}
\looseness=-1
Table~\ref{tab:main} summarizes the main results. \proposed~\textbf{\textit{improves shifter performance across all 21 dynamic backbone–dataset combinations}}, yielding universal improvement regardless of the domain or underlying encoder: up to 114.9\% and 202.5\% improvement on accuracy and F1 score, respectively. Gains are largest on \textbf{GossipCop} and \textbf{Music}, where baseline shifter performance is lowest, and even on \textbf{DBLP-S}, where baselines already achieve moderate shifter performance, \proposed~still yields consistent improvements. Critically, \textbf{\textit{these gains do not come at the expense of stable-node performance}}: accuracy and F1 score on stable nodes are generally preserved or also improved substantially in all cases.

\looseness=-1
\proposed~also improves the static baselines, confirming that the benefit is not contingent on temporal modeling. Notably, EERM, which targets distribution shift through invariant learning, fails to meaningfully improve upon GCN on its own. This reinforces that the shifter problem is fundamentally different from distributional mismatch (See Section~\ref{relatedworks}.2). Dynamic baselines generally outperform their static counterparts, yet even the sequential GCN and EERM, which lack any cross-snapshot information flow, benefit immensely from \proposed~ augmentation.

\begin{wrapfigure}{r}{0.6\textwidth}
  \centering
  \vspace{-20pt}
  \includegraphics[width=0.58\textwidth]{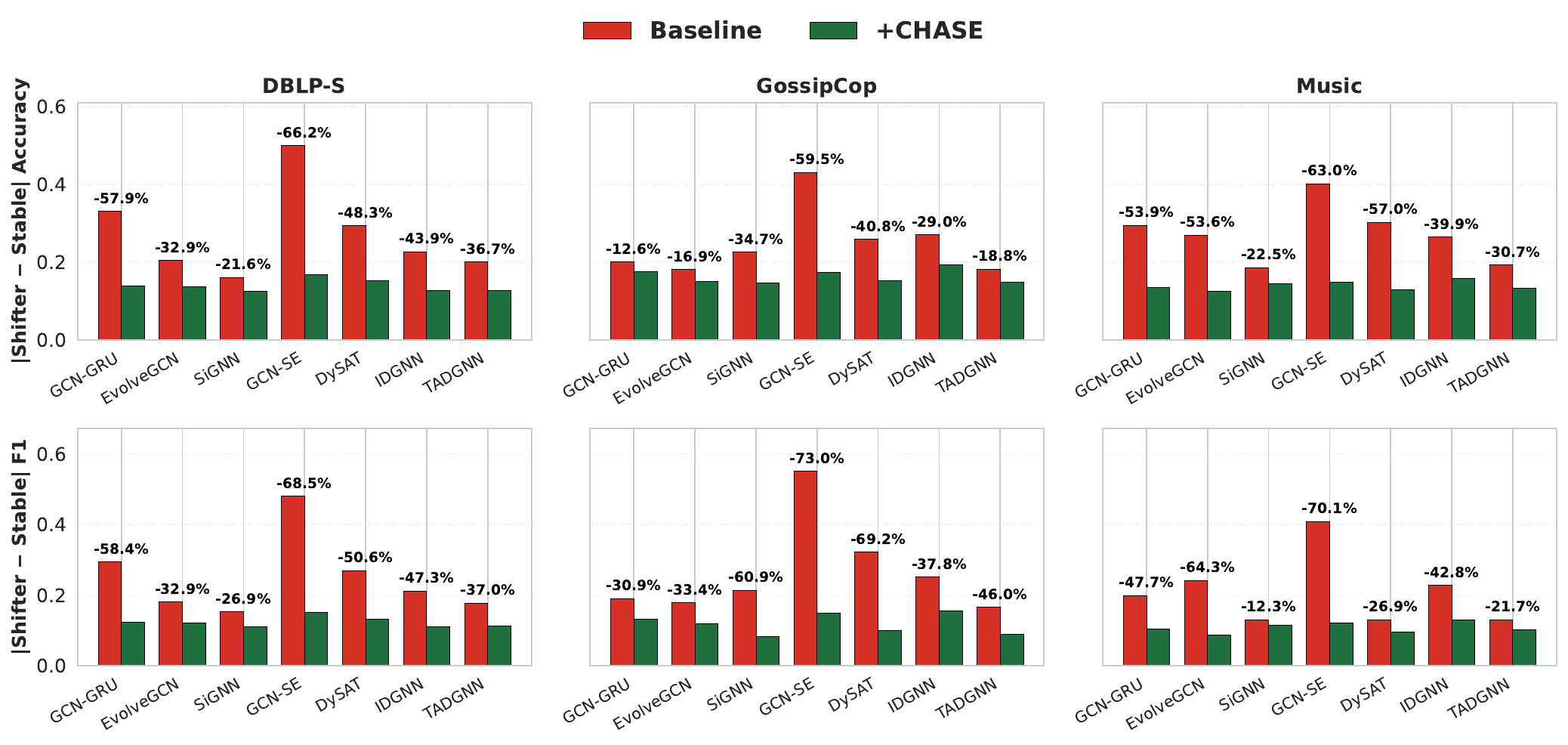}
  \vspace{-10pt}
  \caption{Performance gap between shifters and stable nodes is significantly reduced (up to $-73.0\%$) under \proposed.}
  \label{fig:gap_reduction}
  \vspace{-20pt}
\end{wrapfigure}

\looseness=-1
Meanwhile, the shifter–stable gap narrows substantially under \proposed~across all dynamic backbones and datasets (See Fig.~\ref{fig:gap_reduction}), indicating that our framework \textbf{\textit{reduces the performance disparity between the two node types}}, a desirable property for real-world deployment where reliable classification of both shifters and stable nodes is important.

\begin{table}
\caption{Main results. Shifter (Sh.) and stable node (St.) accuracy (Acc) and F1 score for each backbone, without and \colorbox{green!8}{with \proposed~augmentation.} On shaded rows, \textbf{bold} values indicate improvement over the base, with relative change shown as superscripts. All values are 5-seed means in percentage points; standard deviations are reported in Appendix~\ref{app:results}.}
\label{tab:main}
\centering
\renewcommand{\arraystretch}{1.0}
\resizebox{\textwidth}{!}{%
\begin{tabular}{l llll llll llll}
\toprule
& \multicolumn{4}{c}{\textbf{DBLP-S}} & \multicolumn{4}{c}{\textbf{GossipCop}} & \multicolumn{4}{c}{\textbf{Music}} \\
\cmidrule(lr){2-5} \cmidrule(lr){6-9} \cmidrule(lr){10-13}
\textbf{Model} & Sh.Acc & Sh.F1 & St.Acc & St.F1 & Sh.Acc & Sh.F1 & St.Acc & St.F1 & Sh.Acc & Sh.F1 & St.Acc & St.F1 \\
\midrule
\textbf{GCN} (collapsed) & 73.0 & 76.5 & 96.0 & 97.1 & 48.6 & 22.1 & 69.7 & 40.1 & 38.1 & 23.7 & 64.5 & 22.9 \\
\textbf{GCN} (seq) & 71.8 & 75.5 & 96.1 & 97.1 & 52.8 & 25.0 & 75.6 & 45.3 & 46.3 & 29.0 & 73.6 & 45.0 \\
\rowcolor{green!8} \quad +\proposed~& \textbf{83.2}\textsuperscript{\tiny\textcolor{green!60!black}{\textbf{+15.9\%}}} & \textbf{85.9}\textsuperscript{\tiny\textcolor{green!60!black}{\textbf{+13.8\%}}} & \textbf{96.2}\textsuperscript{\tiny\textcolor{green!60!black}{\textbf{+0.1\%}}} & \textbf{97.2}\textsuperscript{\tiny\textcolor{green!60!black}{\textbf{+0.1\%}}} & \textbf{69.6}\textsuperscript{\tiny\textcolor{green!60!black}{\textbf{+31.8\%}}} & \textbf{60.1}\textsuperscript{\tiny\textcolor{green!60!black}{\textbf{+140.4\%}}} & \textbf{87.7}\textsuperscript{\tiny\textcolor{green!60!black}{\textbf{+16.0\%}}} & \textbf{76.9}\textsuperscript{\tiny\textcolor{green!60!black}{\textbf{+69.8\%}}} & \textbf{80.0}\textsuperscript{\tiny\textcolor{green!60!black}{\textbf{+72.8\%}}} & \textbf{78.6}\textsuperscript{\tiny\textcolor{green!60!black}{\textbf{+171.0\%}}} & \textbf{93.8}\textsuperscript{\tiny\textcolor{green!60!black}{\textbf{+27.4\%}}} & \textbf{89.5}\textsuperscript{\tiny\textcolor{green!60!black}{\textbf{+98.9\%}}} \\
\textbf{EERM} (collapsed) & 73.5 & 77.4 & 96.2 & 97.3 & 47.5 & 20.0 & 69.6 & 44.5 & 39.2 & 22.4 & 64.7 & 25.6 \\
\textbf{EERM} (seq) & 73.6 & 77.3 & 96.3 & 97.3 & 52.5 & 23.4 & 76.1 & 47.1 & 44.0 & 24.6 & 73.8 & 45.4 \\
\rowcolor{green!8} \quad +\proposed~& \textbf{83.1}\textsuperscript{\tiny\textcolor{green!60!black}{\textbf{+12.9\%}}} & \textbf{85.6}\textsuperscript{\tiny\textcolor{green!60!black}{\textbf{+10.7\%}}} & 96.1\textsuperscript{\tiny\textcolor{gray}{\textbf{-0.2\%}}} & 97.2\textsuperscript{\tiny\textcolor{gray}{\textbf{-0.1\%}}} & \textbf{70.0}\textsuperscript{\tiny\textcolor{green!60!black}{\textbf{+33.3\%}}} & \textbf{61.0}\textsuperscript{\tiny\textcolor{green!60!black}{\textbf{+160.7\%}}} & \textbf{87.7}\textsuperscript{\tiny\textcolor{green!60!black}{\textbf{+15.2\%}}} & \textbf{76.9}\textsuperscript{\tiny\textcolor{green!60!black}{\textbf{+63.3\%}}} & \textbf{80.3}\textsuperscript{\tiny\textcolor{green!60!black}{\textbf{+82.5\%}}} & \textbf{79.3}\textsuperscript{\tiny\textcolor{green!60!black}{\textbf{+222.4\%}}} & \textbf{93.9}\textsuperscript{\tiny\textcolor{green!60!black}{\textbf{+27.2\%}}} & \textbf{89.5}\textsuperscript{\tiny\textcolor{green!60!black}{\textbf{+97.1\%}}} \\
\midrule
\textbf{GCN-GRU} & 63.9 & 68.4 & 97.0 & 97.8 & 58.2 & 34.1 & 78.3 & 53.1 & 50.1 & 35.2 & 79.5 & 55.1 \\
\rowcolor{green!8} \quad +\proposed~& \textbf{82.3}\textsuperscript{\tiny\textcolor{green!60!black}{\textbf{+28.8\%}}} & \textbf{85.0}\textsuperscript{\tiny\textcolor{green!60!black}{\textbf{+24.3\%}}} & 96.3\textsuperscript{\tiny\textcolor{gray}{\textbf{-0.8\%}}} & 97.2\textsuperscript{\tiny\textcolor{gray}{\textbf{-0.6\%}}} & \textbf{70.1}\textsuperscript{\tiny\textcolor{green!60!black}{\textbf{+20.4\%}}} & \textbf{64.4}\textsuperscript{\tiny\textcolor{green!60!black}{\textbf{+88.8\%}}} & \textbf{87.7}\textsuperscript{\tiny\textcolor{green!60!black}{\textbf{+12.0\%}}} & \textbf{77.5}\textsuperscript{\tiny\textcolor{green!60!black}{\textbf{+46.0\%}}} & \textbf{80.1}\textsuperscript{\tiny\textcolor{green!60!black}{\textbf{+60.0\%}}} & \textbf{79.2}\textsuperscript{\tiny\textcolor{green!60!black}{\textbf{+124.8\%}}} & \textbf{93.7}\textsuperscript{\tiny\textcolor{green!60!black}{\textbf{+17.8\%}}} & \textbf{89.6}\textsuperscript{\tiny\textcolor{green!60!black}{\textbf{+62.4\%}}} \\
\textbf{EvolveGCN} & 75.7 & 79.2 & 96.2 & 97.2 & 70.0 & 60.1 & 88.1 & 78.0 & 59.9 & 52.2 & 86.9 & 76.3 \\
\rowcolor{green!8} \quad +\proposed~& \textbf{82.4}\textsuperscript{\tiny\textcolor{green!60!black}{\textbf{+8.9\%}}} & \textbf{85.1}\textsuperscript{\tiny\textcolor{green!60!black}{\textbf{+7.5\%}}} & 96.2\textsuperscript{\tiny\textcolor{green!60!black}{\textbf{+0.0\%}}} & 97.2\textsuperscript{\tiny\textcolor{green!60!black}{\textbf{+0.0\%}}} & \textbf{72.3}\textsuperscript{\tiny\textcolor{green!60!black}{\textbf{+3.4\%}}} & \textbf{64.4}\textsuperscript{\tiny\textcolor{green!60!black}{\textbf{+7.1\%}}} & 87.4\textsuperscript{\tiny\textcolor{gray}{\textbf{-0.8\%}}} & 76.3\textsuperscript{\tiny\textcolor{gray}{\textbf{-2.1\%}}} & \textbf{81.0}\textsuperscript{\tiny\textcolor{green!60!black}{\textbf{+35.3\%}}} & \textbf{80.2}\textsuperscript{\tiny\textcolor{green!60!black}{\textbf{+53.7\%}}} & \textbf{93.6}\textsuperscript{\tiny\textcolor{green!60!black}{\textbf{+7.7\%}}} & \textbf{88.8}\textsuperscript{\tiny\textcolor{green!60!black}{\textbf{+16.3\%}}} \\
\textbf{SiGNN} & 71.3 & 75.3 & 87.4 & 90.6 & 49.1 & 28.4 & 71.7 & 49.8 & 71.3 & 68.0 & 89.9 & 81.1 \\
\rowcolor{green!8} \quad +\proposed~& \textbf{83.2}\textsuperscript{\tiny\textcolor{green!60!black}{\textbf{+16.7\%}}} & \textbf{85.8}\textsuperscript{\tiny\textcolor{green!60!black}{\textbf{+13.9\%}}} & \textbf{95.9}\textsuperscript{\tiny\textcolor{green!60!black}{\textbf{+9.7\%}}} & \textbf{97.0}\textsuperscript{\tiny\textcolor{green!60!black}{\textbf{+7.1\%}}} & \textbf{71.3}\textsuperscript{\tiny\textcolor{green!60!black}{\textbf{+45.2\%}}} & \textbf{65.5}\textsuperscript{\tiny\textcolor{green!60!black}{\textbf{+130.6\%}}} & \textbf{86.1}\textsuperscript{\tiny\textcolor{green!60!black}{\textbf{+20.1\%}}} & \textbf{73.9}\textsuperscript{\tiny\textcolor{green!60!black}{\textbf{+48.4\%}}} & \textbf{79.4}\textsuperscript{\tiny\textcolor{green!60!black}{\textbf{+11.4\%}}} & \textbf{78.2}\textsuperscript{\tiny\textcolor{green!60!black}{\textbf{+15.0\%}}} & \textbf{93.8}\textsuperscript{\tiny\textcolor{green!60!black}{\textbf{+4.3\%}}} & \textbf{89.6}\textsuperscript{\tiny\textcolor{green!60!black}{\textbf{+10.5\%}}} \\
\textbf{GCN-SE} & 47.1 & 49.9 & 97.2 & 97.9 & 41.0 & 20.1 & 84.2 & 75.2 & 36.8 & 30.3 & 76.9 & 71.2 \\
\rowcolor{green!8} \quad +\proposed~& \textbf{79.5}\textsuperscript{\tiny\textcolor{green!60!black}{\textbf{+68.8\%}}} & \textbf{82.2}\textsuperscript{\tiny\textcolor{green!60!black}{\textbf{+64.7\%}}} & 96.4\textsuperscript{\tiny\textcolor{gray}{\textbf{-0.8\%}}} & 97.3\textsuperscript{\tiny\textcolor{gray}{\textbf{-0.6\%}}} & \textbf{69.3}\textsuperscript{\tiny\textcolor{green!60!black}{\textbf{+69.0\%}}} & \textbf{60.8}\textsuperscript{\tiny\textcolor{green!60!black}{\textbf{+202.5\%}}} & \textbf{86.8}\textsuperscript{\tiny\textcolor{green!60!black}{\textbf{+3.1\%}}} & \textbf{75.7}\textsuperscript{\tiny\textcolor{green!60!black}{\textbf{+0.7\%}}} & \textbf{79.1}\textsuperscript{\tiny\textcolor{green!60!black}{\textbf{+114.9\%}}} & \textbf{77.3}\textsuperscript{\tiny\textcolor{green!60!black}{\textbf{+155.1\%}}} & \textbf{94.0}\textsuperscript{\tiny\textcolor{green!60!black}{\textbf{+22.2\%}}} & \textbf{89.5}\textsuperscript{\tiny\textcolor{green!60!black}{\textbf{+25.7\%}}} \\
\textbf{DySAT} & 61.2 & 67.1 & 90.7 & 94.0 & 49.8 & 29.0 & 75.8 & 61.2 & 39.9 & 34.8 & 70.2 & 47.9 \\
\rowcolor{green!8} \quad +\proposed~& \textbf{81.2}\textsuperscript{\tiny\textcolor{green!60!black}{\textbf{+32.7\%}}} & \textbf{84.1}\textsuperscript{\tiny\textcolor{green!60!black}{\textbf{+25.3\%}}} & \textbf{96.5}\textsuperscript{\tiny\textcolor{green!60!black}{\textbf{+6.4\%}}} & \textbf{97.4}\textsuperscript{\tiny\textcolor{green!60!black}{\textbf{+3.6\%}}} & \textbf{69.4}\textsuperscript{\tiny\textcolor{green!60!black}{\textbf{+39.4\%}}} & \textbf{60.6}\textsuperscript{\tiny\textcolor{green!60!black}{\textbf{+109.0\%}}} & \textbf{84.7}\textsuperscript{\tiny\textcolor{green!60!black}{\textbf{+11.7\%}}} & \textbf{70.5}\textsuperscript{\tiny\textcolor{green!60!black}{\textbf{+15.2\%}}} & \textbf{80.3}\textsuperscript{\tiny\textcolor{green!60!black}{\textbf{+101.3\%}}} & \textbf{79.3}\textsuperscript{\tiny\textcolor{green!60!black}{\textbf{+127.9\%}}} & \textbf{93.4}\textsuperscript{\tiny\textcolor{green!60!black}{\textbf{+33.0\%}}} & \textbf{88.8}\textsuperscript{\tiny\textcolor{green!60!black}{\textbf{+85.4\%}}} \\
\textbf{IDGNN} & 73.3 & 75.9 & 96.0 & 97.1 & 58.0 & 46.3 & 85.1 & 71.4 & 67.4 & 66.1 & 93.9 & 88.9 \\
\rowcolor{green!8} \quad +\proposed~& \textbf{83.3}\textsuperscript{\tiny\textcolor{green!60!black}{\textbf{+13.6\%}}} & \textbf{86.0}\textsuperscript{\tiny\textcolor{green!60!black}{\textbf{+13.2\%}}} & \textbf{96.0}\textsuperscript{\tiny\textcolor{green!60!black}{\textbf{+0.0\%}}} & \textbf{97.1}\textsuperscript{\tiny\textcolor{green!60!black}{\textbf{+0.0\%}}} & \textbf{66.9}\textsuperscript{\tiny\textcolor{green!60!black}{\textbf{+15.4\%}}} & \textbf{58.2}\textsuperscript{\tiny\textcolor{green!60!black}{\textbf{+25.8\%}}} & \textbf{86.1}\textsuperscript{\tiny\textcolor{green!60!black}{\textbf{+1.3\%}}} & \textbf{73.9}\textsuperscript{\tiny\textcolor{green!60!black}{\textbf{+3.4\%}}} & \textbf{77.8}\textsuperscript{\tiny\textcolor{green!60!black}{\textbf{+15.4\%}}} & \textbf{76.0}\textsuperscript{\tiny\textcolor{green!60!black}{\textbf{+15.0\%}}} & 93.7\textsuperscript{\tiny\textcolor{gray}{\textbf{-0.2\%}}} & \textbf{89.1}\textsuperscript{\tiny\textcolor{green!60!black}{\textbf{+0.1\%}}} \\
\textbf{TADGNN} & 76.6 & 79.9 & 96.7 & 97.6 & 65.9 & 50.6 & 84.1 & 67.2 & 69.1 & 62.1 & 88.4 & 75.1 \\
\rowcolor{green!8} \quad +\proposed~& \textbf{83.2}\textsuperscript{\tiny\textcolor{green!60!black}{\textbf{+8.7\%}}} & \textbf{85.9}\textsuperscript{\tiny\textcolor{green!60!black}{\textbf{+7.5\%}}} & 96.0\textsuperscript{\tiny\textcolor{gray}{\textbf{-0.8\%}}} & 97.0\textsuperscript{\tiny\textcolor{gray}{\textbf{-0.6\%}}} & \textbf{69.8}\textsuperscript{\tiny\textcolor{green!60!black}{\textbf{+5.9\%}}} & \textbf{61.4}\textsuperscript{\tiny\textcolor{green!60!black}{\textbf{+21.3\%}}} & \textbf{84.6}\textsuperscript{\tiny\textcolor{green!60!black}{\textbf{+0.6\%}}} & \textbf{70.4}\textsuperscript{\tiny\textcolor{green!60!black}{\textbf{+4.7\%}}} & \textbf{80.2}\textsuperscript{\tiny\textcolor{green!60!black}{\textbf{+16.1\%}}} & \textbf{79.0}\textsuperscript{\tiny\textcolor{green!60!black}{\textbf{+27.2\%}}} & \textbf{93.6}\textsuperscript{\tiny\textcolor{green!60!black}{\textbf{+5.9\%}}} & \textbf{89.2}\textsuperscript{\tiny\textcolor{green!60!black}{\textbf{+18.7\%}}} \\
\bottomrule
\end{tabular}
}
\vspace{-3ex}
\end{table}

\vspace{-1ex}
\subsection{Ablation Study}
\vspace{-1ex}
\begin{wraptable}[7]{r}{0.55\textwidth}
\vspace{-25pt}
\caption{Ablation study on GCN-GRU.}
\label{tab:ablation}
\centering
\resizebox{0.55\textwidth}{!}{%
\begin{tabular}{l ll ll ll}
\toprule
& \multicolumn{2}{c}{\textbf{DBLP-S}} & \multicolumn{2}{c}{\textbf{GossipCop}} & \multicolumn{2}{c}{\textbf{Music}} \\
\cmidrule(lr){2-3} \cmidrule(lr){4-5} \cmidrule(lr){6-7}
\textbf{Config} & Sh.Acc & Sh.F1 & Sh.Acc & Sh.F1 & Sh.Acc & Sh.F1 \\
\midrule
Baseline & 63.9$_{\pm 1.0}$ & 68.4$_{\pm 1.1}$ & 58.2$_{\pm 4.7}$ & 34.1$_{\pm 12.4}$ & 50.1$_{\pm 7.0}$ & 35.2$_{\pm 13.6}$ \\
+Contr. & 65.9$_{\pm 0.6}$ & 69.9$_{\pm 0.6}$ & 60.4$_{\pm 4.8}$ & 45.5$_{\pm 3.4}$ & 53.6$_{\pm 7.0}$ & 44.6$_{\pm 7.4}$ \\
+Hom.wt. & 68.2$_{\pm 1.0}$ & 71.9$_{\pm 1.1}$ & 61.9$_{\pm 4.5}$ & 48.7$_{\pm 3.7}$ & 58.1$_{\pm 4.6}$ & 51.7$_{\pm 4.5}$ \\
+Cls. & 70.7$_{\pm 1.1}$ & 73.3$_{\pm 1.8}$ & 63.4$_{\pm 3.6}$ & 51.3$_{\pm 3.6}$ & 62.2$_{\pm 4.2}$ & 56.7$_{\pm 3.6}$ \\
+Detect. & 76.6$_{\pm 1.8}$ & 80.2$_{\pm 2.2}$ & 68.4$_{\pm 1.4}$ & 59.2$_{\pm 1.4}$ & 76.5$_{\pm 1.0}$ & 75.3$_{\pm 1.0}$ \\
\rowcolor{green!8} \textbf{Full} & \textbf{82.3}$_{\pm 0.5}$ & \textbf{85.0}$_{\pm 0.6}$ & \textbf{70.1}$_{\pm 0.9}$ & \textbf{64.4}$_{\pm 1.5}$ & \textbf{80.1}$_{\pm 1.0}$ & \textbf{79.2}$_{\pm 1.0}$ \\
\bottomrule
\end{tabular}
}
\end{wraptable}

To isolate the contribution of each \proposed~component, we incrementally add components to the GCN-GRU baseline across all three datasets (See Table~\ref{tab:ablation}). Starting from the baseline MLP classifier on $[\mathbf{h}_v^{(t)}]$, we first add the contrastive loss $\mathcal{L}_{\mathrm{hom}}$ (+Contr.), and then its $(1 - h_{\mathrm{curr}})$ weighting (+Hom.wt.). We next augment the classifier input with $\mathbf{z}_{\mathrm{self},v}^{(t)}$ and $\hat{\mathbf{y}}_v^{(t{-}1)}$ (+Cls.), then add the change detection head (and $\mathcal{L}_{\mathrm{change}}$) whose output $p_{\mathrm{change}}$ is fed into the classifier (+Detect.), and finally add sqrt-dampened weighting to $\mathcal{L}_{\mathrm{cls}}$ (Full).

Every component contributes monotonically to shifter performance. Particularly, \textbf{\textit{the largest single jump occurs at +Detect.}}, confirming the analysis in Section~\ref{analysis}: the preceding steps give the classifier the \textit{capacity} to deviate from old predictions (via $\mathbf{z}_{\mathrm{self},v}^{(t)}$ and $\hat{\mathbf{y}}_v^{(t{-}1)}$), but without $p_{\mathrm{change}}$ it lacks the \textit{signal} for when to do so. We additionally compare against a rule-based variant in Appendix~\ref{app:further}.1.

\begin{figure}[t]
\centering
\includegraphics[width=0.9\linewidth]{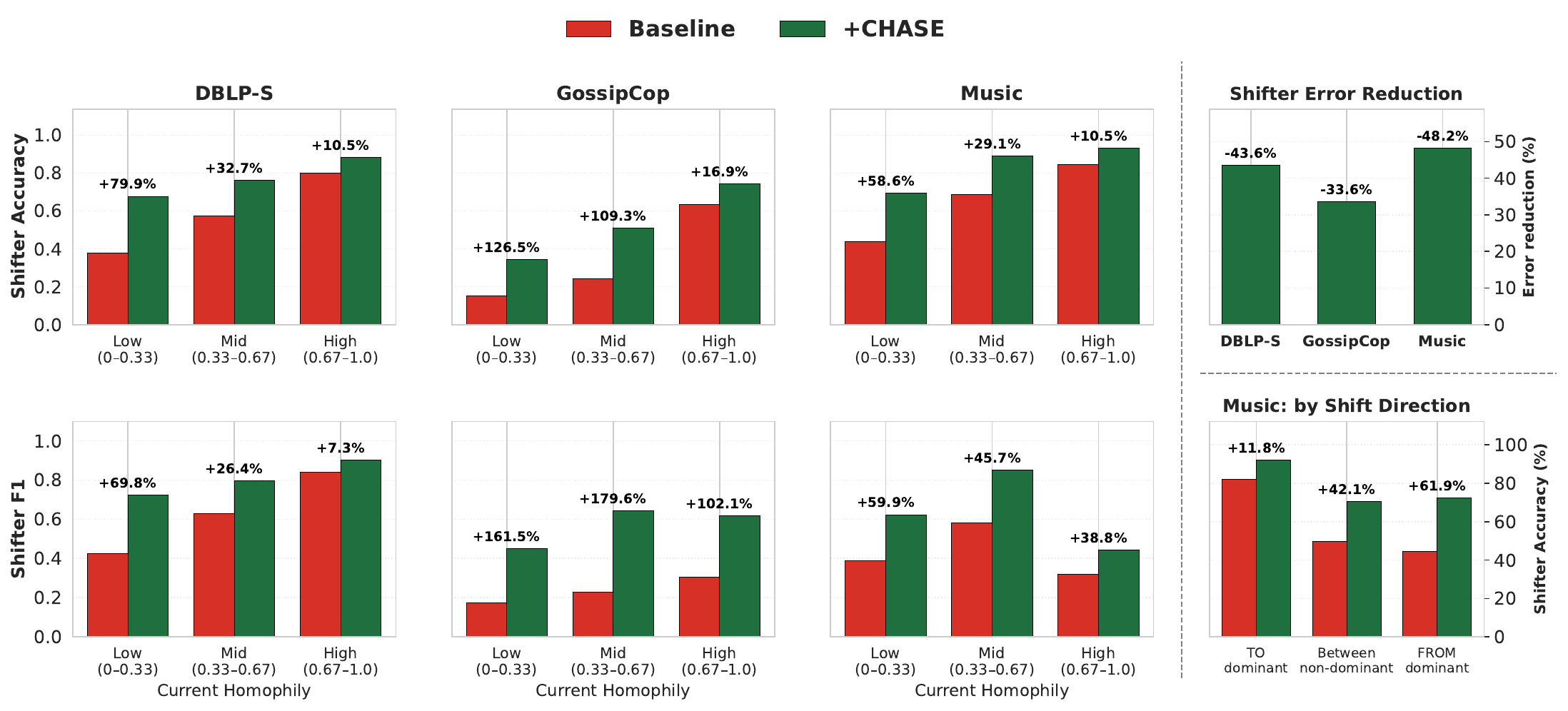}
\vspace{-2ex}
\caption{Performance improvements after \proposed~augmentation, averaged across all models. (Left) Accuracy and F1 score gains are greater for nodes with lower current homophily. (Top right) Classification error on shifters is substantially reduced. (Bottom right) On the multiclass \textbf{Music} dataset, shifts \textit{away from} the dominant class benefit much more than shifts \textit{toward} it.}
\vspace{-3ex}
\label{fig:alignment}
\end{figure}

\subsection{Further Analysis}
\vspace{-1ex}
\noindent{\textbf{\proposed~improvement aligns with homophily.}}
The analysis in Section~\ref{analysis} identifies low current homophily as the structural condition under which embedding inertia is most severe. If \proposed~correctly addresses this mechanism, its gains should be concentrated more on these nodes. Fig.~\ref{fig:alignment}(Left) confirms this: across all backbones and datasets, \proposed's relative improvement on shifter accuracy and F1 is largest for nodes with the lowest current homophily and decreases monotonically as homophily increases. This alignment between the diagnostic finding and the empirical improvement validates the analysis-driven design of \proposed.

\looseness=-1
\noindent{\textbf{Error reduction toward the correct label.}}
A natural concern is whether \proposed~simply learns to avoid predicting the old label, rather than actively predicting the correct new one. On the binary datasets, these are equivalent, i.e., any move away from the old label is toward the correct one. On the multiclass \textbf{Music} dataset (5 classes), however, a model that merely suppresses the old-label prediction would still misclassify 75\% of the time among the remaining classes. Fig.~\ref{fig:alignment}(Top right) shows that \proposed~achieves substantial shifter error reduction on all three datasets, with greatest improvement on \textbf{Music}, confirming that the improvement reflects genuine classification rather than mere bias removal.

\looseness=-1
\noindent{\textbf{Larger gains on shifts away from the dominant class.}}
On the \textbf{Music} dataset, class 0 (Rock genre) accounts for the majority of class labels. Under standard cross-entropy training, the loss is dominated by the dominant class, biasing learned representations in its favor. As a result, shifts \textit{toward} the dominant class are naturally easier for baselines, while shifts \textit{away from} it are harder. Fig.~\ref{fig:alignment}(Bottom right) confirms this: baseline accuracy is lowest for shifts away from the dominant class, moderate for shifts between non-dominant classes, and highest for shifts toward it. \proposed's performance gain follows the opposite ordering: largest for shifts away, moderate between non-dominant classes, and smallest toward. This indicates that \proposed~provides the most benefit where embedding inertia is strongest.

\begin{table}[t]
\centering
\small
\caption{Shifter accuracy by baseline inertia ($\Delta_c$), pooled over all backbones and datasets.}
\vspace{-1ex}
\label{tab:inertia}
\begin{tabular}{lcc}
\toprule
Baseline $\Delta_c$ & Baseline Acc.\ (\%) & Gain from +\proposed~(\%p) \\
\midrule
$< -0.25$ (most inertial) & 3.0 & +40.1 \\
$-0.25$ to $-0.05$ & 15.9 & +37.5 \\
$-0.05$ to $0.05$ & 61.1 & +17.9 \\
$> 0.05$ (non-inertial) & 94.0 & 0.0 \\
\bottomrule
\end{tabular}
\vspace{-3ex}
\end{table}

\looseness=-1
\noindent{\textbf{\proposed~directly repairs embedding inertia.}}
We further examine embedding inertia across all shifters, beyond specific structural or class conditions. First, we bin shifters by their centroid displacement $\Delta_c$ before applying \proposed~(See Table~\ref{tab:inertia}). Baseline accuracy drops sharply as inertia increases, and \proposed's gain is concentrated on the inertial bins (up to +40.1\%p), with the smallest gain in the near-zero bin and none for non-inertial shifters. Second, we compare each shifter's $\Delta_c$ before and after adding \proposed~to the same model. Shifters corrected by \proposed~move from the old-class side to the new-class side ($\Delta_c$: $-0.19 \to +0.09$), while already-correct shifters remain on the new-class side ($+0.27 \to +0.19$). Together, these results complement our analysis on embedding inertia driving shifter errors and show that \proposed~improves performance by reversing it.

Additional analyses are provided in Appendix~\ref{app:further}, including further exploration on $p_{\mathrm{change}}$, improvement stratified by number of shifts per node, per-snapshot consistency with comparison against retraining baselines that have access to ground-truth labels from previous test snapshots, robustness to incorrect $\hat{\mathbf{y}}^{(t{-}1)}$ and input noise, hyperparameter sensitivity, case studies, and computation overhead.

\vspace{-1ex}
\section{Conclusion}
\vspace{-1ex}
In this work, we identified a systematic failure mode shared by all major dynamic GNN architectures: \textit{shifters}, nodes whose labels change between snapshots, are classified with substantially lower performance than stable nodes, with errors concentrated on the old label. We traced this to \textit{embedding inertia}, where neighborhood aggregation anchors representations to the old class, and proposed \textbf{\proposed}, a model-agnostic framework that counteracts this effect through a homophily-guided contrastive loss, a change detection head, and a change-aware classifier. \proposed~improves shifter performance across all tested backbone–dataset combinations while preserving stable-node performance. We additionally contributed three dynamic graph benchmarks with naturally shifting labels, addressing a gap in existing evaluation resources.

\vspace{-1ex}
\section{Limitations and Future Work}
\vspace{-1ex}
\proposed~treats label change as binary: a node either shifted or it did not. In practice, many transitions are gradual, e.g., a user's consumption pattern may drift over several snapshots before crossing the labeling threshold, yet our current setup treats the node as stable until the threshold is crossed. This means the contrastive loss and detection head do not act on nodes approaching the decision boundary, potentially allowing embedding inertia to accumulate before the shift is registered. A \textit{continuous transition intensity score} could allow \proposed~to recognize and act on nodes that are mid-transition, rather than ignoring them until the label flips.

Such a formulation would also support an \textit{anticipatory} extension: predicting which nodes will shift before the transition occurs, rather than correcting after the fact. \proposed's structural signals, e.g., declining current homophily and rising neighborhood drift, are plausible early indicators, but the current formulation targets detection at time $t$ rather than prediction at $t{+}1$. Combining transition intensity scores with forward-looking detection is a natural next step toward early-warning systems for label change in dynamic graphs.

Finally, \proposed~currently operates on discrete snapshot sequences, which fundamentally limits both the granularity of transition intensity measurement and the temporal resolution of anticipatory detection. Extending \proposed~to continuous-time dynamic graph models such as TGN~\cite{tgn} and DyRep~\cite{dyrep}, where events are observed as they occur, is a promising next step. In this setting, drift signals could be computed over sliding time windows of recent interactions rather than between fixed snapshots, the homophily proxy could use each neighbor's most recent prediction rather than a snapshot-level aggregate, and the change detection head could output $p_{\mathrm{change}}$ per event rather than per snapshot, enabling finer-grained tracking of gradual transitions and more timely shift prediction.

\vspace{-1ex}
\section*{Acknowledgements}
\vspace{-1ex}
This work was supported by the National Research Foundation of Korea (NRF) grants funded by the Korea government (MSIT) (RS-2024-00406985) and the Ministry of Science and ICT (RS-2022-NR068758), and the Institute of Information \& Communications Technology Planning \& Evaluation (IITP) grant funded by the Korea government (MSIT) (RS-2022-II220157).

\clearpage
\bibliographystyle{plainnat}
\bibliography{reference}

\clearpage
\appendix

\rule[0pt]{\columnwidth}{4pt}
\begin{center}
    \Large{\bf{Supplementary Material for \\[0.5em]
    Chasing Label Shifters: A Change-Aware Framework \\
    for Dynamic Graph Node Classification}}
\end{center}
\vspace{2ex}

\DoToC

\clearpage

\section{Dataset Construction Details} \label{app:datasets}
We describe the construction of each dataset in detail, including data sources, node and edge definitions, label derivation, and feature extraction. Dataset statistics are summarized in Table~\ref{tab:datasets} in the main text.

\noindent{\textbf{DBLP-S.}}
\begin{itemize}[leftmargin = 2mm]
    \item \textbf{Source.} DBLP-v11 citation network dataset\footnote{\url{https://dblp.org/}}. Our dataset is derived from GCN-SE's~\cite{gcnse} DBLP-E, which was introduced as a label-changing dataset but lacked node features and used a fixed node pool across snapshots. We extend it with text-based node features and an evolving node pool.
    \item \textbf{Node set.} All authors who have at least one paper corresponding to one of the two label categories in the final snapshot. The node set evolves across snapshots: authors enter the graph when they first publish a relevant paper, and inactive authors are retained but become isolated. The dataset spans 14 yearly snapshots (2005--2018) with 34,717 unique nodes across all snapshots.
    \item \textbf{Edge construction.} An edge $(u, v)$ exists in snapshot $G^{(t)}$ if authors $u$ and $v$ co-authored at least one paper in year $t$, weighted by the number of shared papers.
    \item \textbf{Label derivation.} Each author is labeled based on the \texttt{fos.name} (field of study) metadata of their papers: 0 for Computer Networks and 1 for Machine Learning. An author's label at snapshot $G^{(t)}$ is determined by whichever field accounts for the majority of their publications in that year.
    \item \textbf{Node features.} 768-dimensional feature vectors are obtained by extracting the [CLS] token from a pretrained BERT~\cite{bert} model (\texttt{bert-base-uncased}) applied to the concatenated title and abstract of each author's papers (corresponding to one of the two label categories) in that year.
\end{itemize}

\noindent{\textbf{GossipCop.}}
\begin{itemize}[leftmargin = 2mm]
    \item \textbf{Source.} FakeNewsNet~\cite{fakenewsnet} containing user interactions with real or fabricated news articles on celebrity gossip, constructed by combining the datasets from DECOR~\cite{decor} and UPFD~\cite{upfd}.
    \item \textbf{Node set.} Users who engaged with (retweeted) at least 3 articles in the dataset. The dataset spans 37 weekly snapshots (November 2017 -- July 2018) with 6,091 unique nodes across all snapshots.
    \item \textbf{Edge construction.} An edge $(u, v)$ exists in snapshot $G^{(t)}$ if users $u$ and $v$ both engaged with the same article within week $t$, weighted by the number of co-consumed articles.
    \item \textbf{Label derivation.} A user is labeled $1$ (fake-news consumer) at snapshot $G^{(t)}$ if more than 50\% of the articles they engaged with in that week are fabricated, and $0$ otherwise.
    \item \textbf{Node features.} Each user's 768-dimensional feature vector is the mean of the embeddings of all articles they engaged with in that snapshot, where article embeddings are the [CLS] token from \texttt{bert-base-uncased} applied to the article text.
\end{itemize}

\noindent{\textbf{Music.}}
\begin{itemize}[leftmargin = 2mm]
    \item \textbf{Source.} Constructed from the LastFM-song-listens~\cite{lastfm1, lastfm2} dataset and the Million Song Dataset~\cite{millionsong}.
    \item \textbf{Node set.} All listeners active within the collection period. The dataset spans 24 monthly snapshots (January 2007 -- December 2008) with 878 unique nodes across all snapshots.
    \item \textbf{Edge construction.} An edge $(u, v)$ exists in snapshot $G^{(t)}$ if users $u$ and $v$ both listened to the same track within month $t$, weighted by the number of co-listened tracks.
    \item \textbf{Label derivation.} Each listener is labeled by their dominant genre in that month, determined by the most common genre across all tracks listened to in that month. The genre taxonomy consists of 5 classes: Rock (0), Metal (1), Electronic (2), Pop/Urban (3), and Niche/Diverse (4).
    \looseness=-1
    \item \textbf{Node features.} 779-dimensional feature vectors composed of four groups: (i) 768-dimensional semantic features, obtained by averaging the [CLS] token from \texttt{bert-base-uncased} over all tracks' user-contributed tags, capturing the mood and style descriptors of the user's monthly listening; (ii) audio properties (duration, loudness, mode, tempo), averaged over all tracks listened to in that month; (iii) popularity signals (artist familiarity, artist hotness, song hotness), averaged over all tracks listened to in that month, capturing mainstream vs.\ niche listening; and (iv) behavioral features computed from all plays: log play count, artist diversity (unique artist count), repeat ratio (fraction of plays that are re-listens), and temporal spread (standard deviation of listening timestamps).
\end{itemize}

\section{Extended Diagnostic Analysis} \label{app:analysis}
\begin{figure}[!h]
\centering
\includegraphics[width=0.95\linewidth]{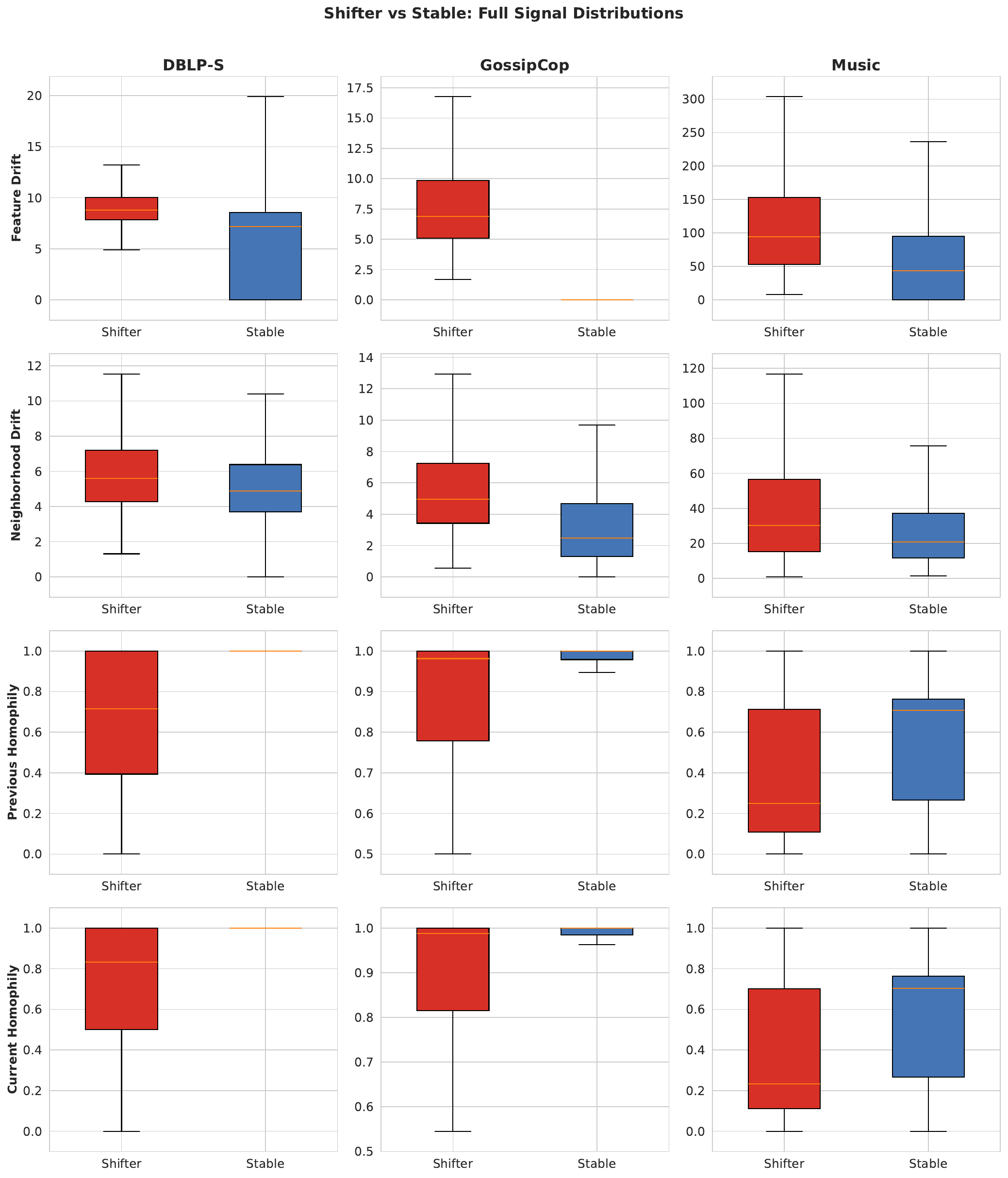}
\caption{All four signals distinguish shifters from stable nodes, consistently across all three datasets.}
\label{fig:app_signals1}
\end{figure}

\begin{figure}[p]
\centering
\includegraphics[width=0.95\linewidth]{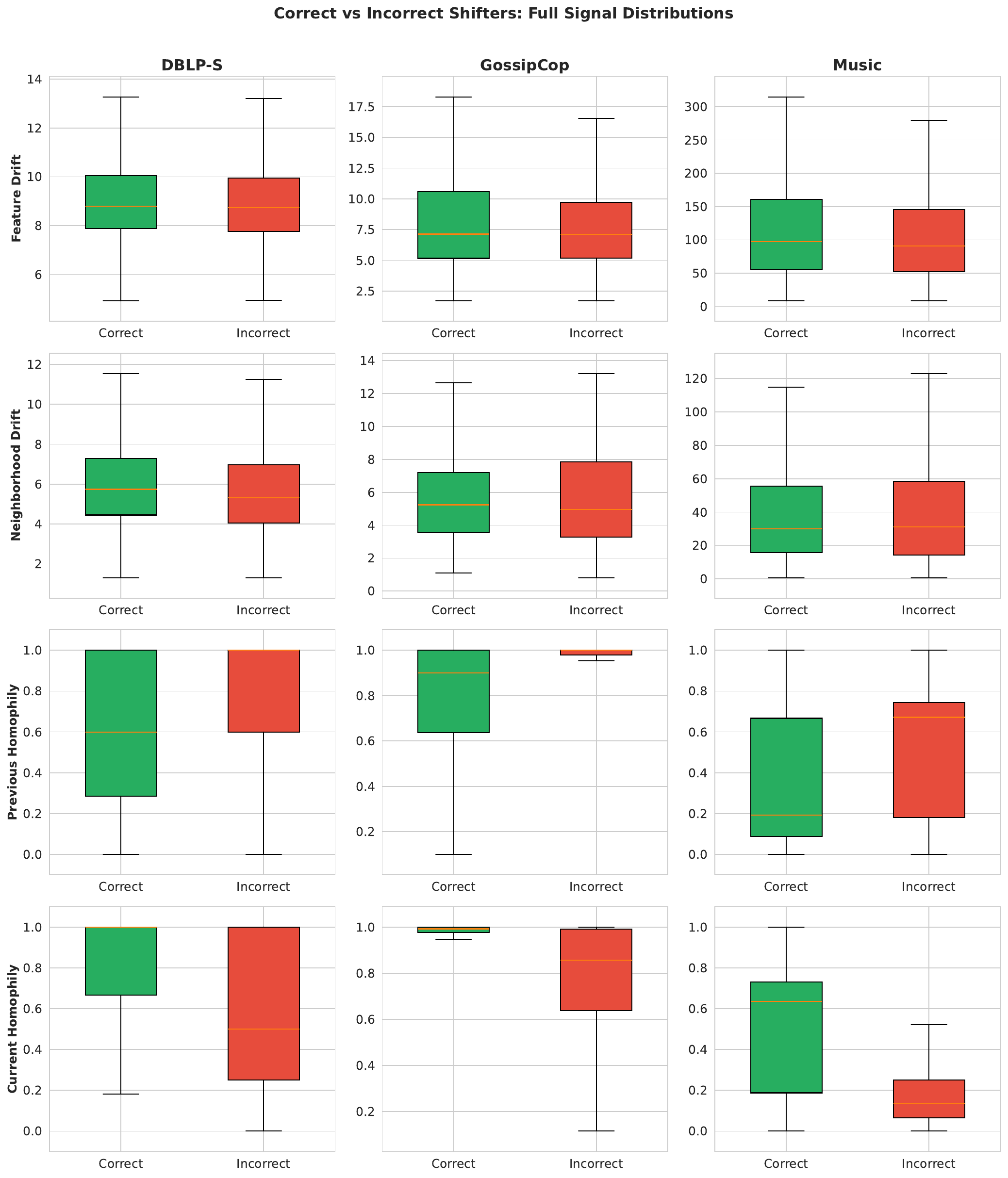}
\caption{Homophily signals meaningfully separate correctly from incorrectly classified shifters while the drift signals lose their discriminative power, consistently across all three datasets.}
\label{fig:app_signals2}
\end{figure}

\begin{figure}[p]
\centering
\includegraphics[width=\linewidth]{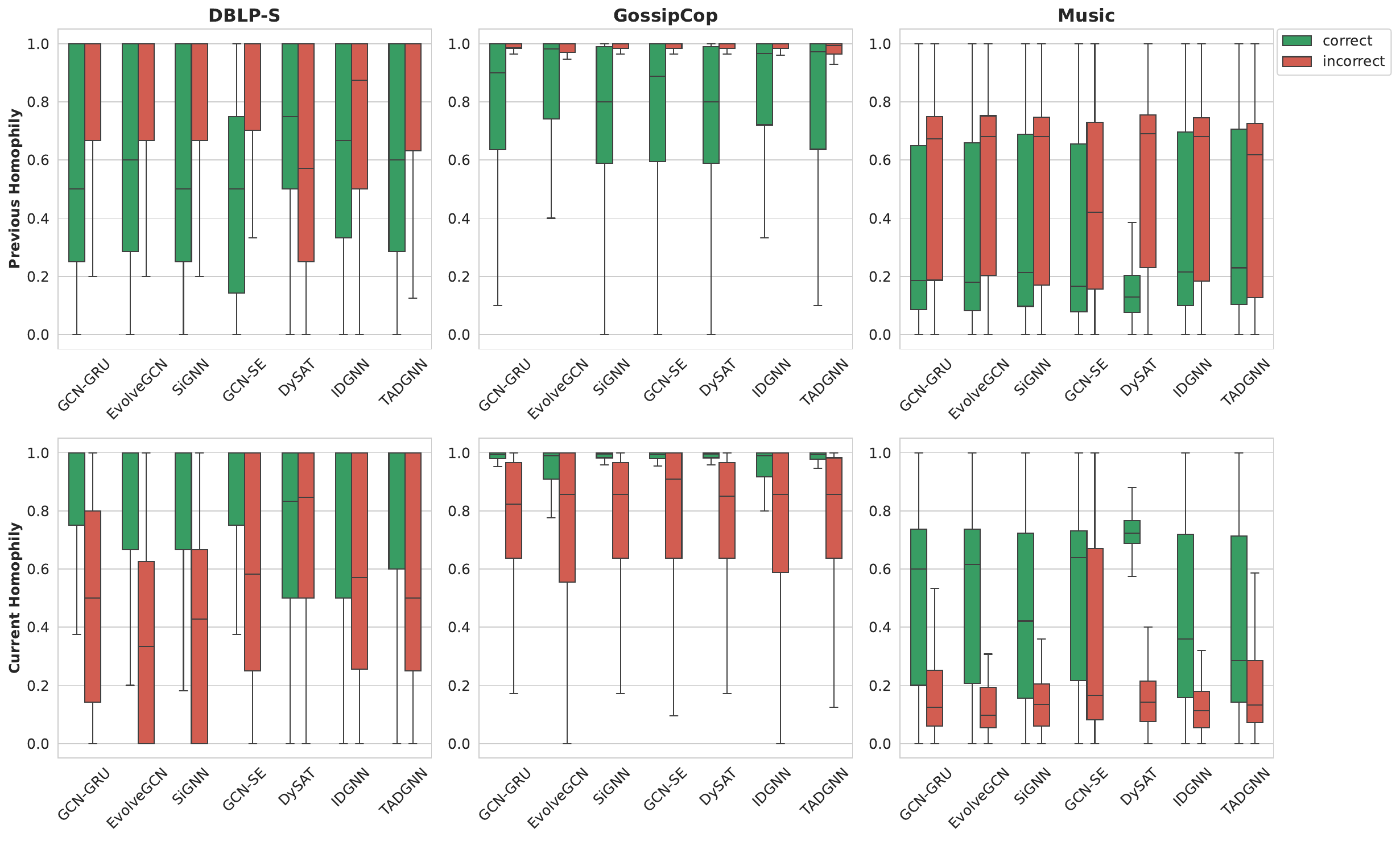}
\caption{The homophily patterns (high current homophily and low previous homophily for correctly classified shifters, vice versa for incorrectly classified shifters) are consistent across all models and datasets.}
\label{fig:app_homophily}
\end{figure}

\begin{figure}[p]
\centering
\includegraphics[width=\linewidth]{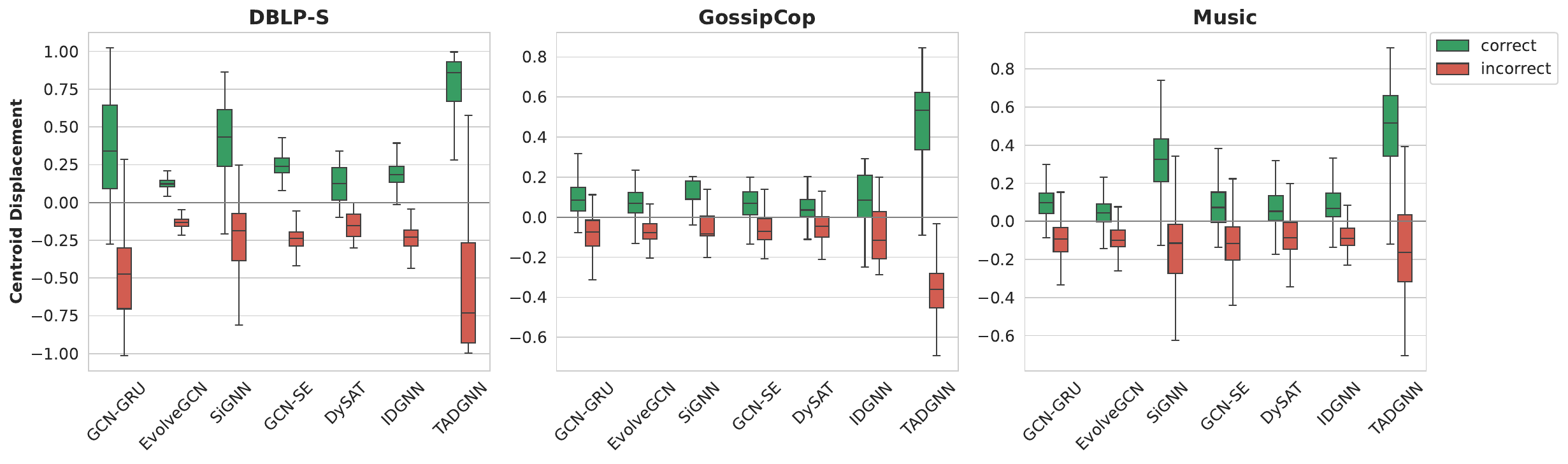}
\caption{Correctly classified shifters generally show positive centroid displacement and incorrectly classified shifters generally display negative centroid displacement, consistently across all models and datasets.}
\label{fig:app_cd}
\end{figure}

Figs.~\ref{fig:app_signals1}--~\ref{fig:app_cd} contain per-model and per-dataset breakdowns of our analysis in Section~\ref{analysis}, demonstrating that our findings are universal and not limited to specific cases.

\clearpage
\clearpage

\clearpage

\clearpage

\vspace{-2ex}
\section{Pseudocode and Computational Complexity} \label{app:method}
\vspace{-1ex}
\subsection{Pseudocode}
Algorithm~\ref{alg:chase} summarizes the \proposed~forward pass for a single snapshot $G^{(t)}$.

\begin{algorithm}[t]
\caption{\proposed~forward pass at snapshot $G^{(t)}$}
\label{alg:chase}
\begin{algorithmic}[1]
\REQUIRE Graph $\mathcal{G}^{(t)}=(V^{(t)},E^{(t)},X^{(t)},y^{(t)})$, previous node features $X^{(t-1)}$, previous edges $E^{(t{-}1)}$, previous hidden state $\mathbf{h}^{(t{-}1)}$, previous predictions $\hat{\mathbf{y}}^{(t{-}1)}$
\ENSURE Node predictions $\hat{\mathbf{y}}^{(t)}$, updated hidden state $\mathbf{h}^{(t)}$, change probabilities $\mathbf{p}_{\mathrm{change}}^{(t)}$
\STATE $\mathbf{h}^{(t)} \gets \mathrm{BaseEncoder}(\mathbf{X}^{(t)}, E^{(t)}, \mathbf{h}^{(t{-}1)})$ \COMMENT{Any dynamic GNN — unmodified}
\STATE $\mathbf{z}_{\mathrm{self}}^{(t)} \gets \mathrm{ReLU}(\mathbf{W}_{\mathrm{self}} \, \mathbf{X}^{(t)})$ \COMMENT{Aggregation-free self-projection}
\FOR{each node $v \in V^{(t)}$}
  \STATE $\mathrm{self\_drift}_v^{(t)} \gets \|\mathbf{x}_v^{(t)} - \mathbf{x}_v^{(t{-}1)}\|$
  \STATE $\mathrm{nbr\_drift}_v^{(t)} \gets \|\bar{\mathbf{x}}_{\mathcal{N}_v}^{(t)} - \bar{\mathbf{x}}_{\mathcal{N}_v}^{(t{-}1)}\|$
  \STATE $h_{\mathrm{proxy}}(v, t) \gets \mathrm{mean}_{u \in \mathcal{N}_v^{(t{-}1)}} \hat{y}_u^{(t{-}1)}[\arg\max \hat{y}_v^{(t{-}1)}]$
\ENDFOR
\STATE $\mathbf{p}_{\mathrm{change}}^{(t)} \gets \sigma\!\big(\mathrm{MLP}_{\mathrm{detect}}([\mathbf{h}^{(t)};\, \mathrm{self\_drift}^{(t)};\, \mathrm{nbr\_drift}^{(t)};\, h_{\mathrm{proxy}}^{(t)}])\big)$
\STATE $\hat{\mathbf{y}}^{(t)} \gets \mathrm{MLP}_{\mathrm{cls}}([\mathbf{h}^{(t)};\, \mathbf{z}_{\mathrm{self}}^{(t)};\, \mathbf{p}_{\mathrm{change}}^{(t)};\, \hat{\mathbf{y}}^{(t{-}1)}])$
\IF{training}
  \STATE Compute $\mathcal{L}_{\mathrm{cls}}$, $\mathcal{L}_{\mathrm{hom}}$ (over shifters, weighted by $1 - h_{\mathrm{curr}}$), $\mathcal{L}_{\mathrm{change}}$
  \STATE $\mathcal{L} \gets \mathcal{L}_{\mathrm{cls}} + \lambda_{\mathrm{hom}} \cdot \mathcal{L}_{\mathrm{hom}} + \lambda_{\mathrm{change}} \cdot \mathcal{L}_{\mathrm{change}}$
\ENDIF
\RETURN $\hat{\mathbf{y}}^{(t)}, \mathbf{h}^{(t)}, \mathbf{p}_{\mathrm{change}}^{(t)}$
\end{algorithmic}
\end{algorithm}

In practice, the feature drift and neighborhood drift signals (lines 4–5) and the current homophily weights used in $\mathcal{L}_{\mathrm{hom}}$ are precomputed once per training run over all snapshots; they are shown inline for clarity. The proxy for previous homophily (line 6) depends on the model's predictions and is computed at each forward pass.

\vspace{-1ex}
\subsection{Computational Complexity}
Let $n = |V^{(t)}|$ denote the number of nodes, $e = |E^{(t)}|$ the number of edges, $d$ the feature dimension, $C$ the number of classes, and $d_h$ the hidden dimension size.

\noindent{\textbf{Base encoder.}}
The cost of the base encoder is unchanged by \proposed.

\noindent{\textbf{Self-projection.}}
A single linear layer: $O(n \cdot d \cdot d_h)$.

\noindent{\textbf{Structural signals (precomputed once).}}
Feature drift requires $O(n \cdot d)$ per snapshot. Neighborhood drift requires computing mean neighbor features at $t$ and $t{-}1$, each $O(e \cdot d)$, followed by an $O(n \cdot d)$ difference. Current homophily weights require $O(e \cdot C)$ per snapshot. These are computed once over all training snapshots before training begins.

\noindent{\textbf{Homophily proxy (per forward pass).}}
Requires one gather and a scatter-mean over previous edges: $O(|E^{(t{-}1)}| \cdot C)$ per snapshot.

\noindent{\textbf{Change detection head.}}
An MLP over $n$ nodes with input dimension $d_h + 3$: $O(n \cdot d_h)$.

\noindent{\textbf{Change-aware classifier.}}
An MLP over $n$ nodes with input dimension $d_h + d_h + 1 + C$: $O(n \cdot (d_h + C) \cdot d_h)$.

\noindent{\textbf{Contrastive loss (training only).}}
Computing centroids from stable nodes: $O(n \cdot d_h)$. Per-shifter cosine similarity: $O(n_{\mathrm{shifter}}^{(t)} \cdot d_h)$. Since $n_{\mathrm{shifter}}^{(t)} \ll n$ in practice, this is negligible.

\noindent{\textbf{Overall.}}
The \proposed~overhead is $O(n \cdot d \cdot d_h + |E^{(t{-}1)}| \cdot C)$ per snapshot, dominated by the self-projection and homophily proxy. The one-time precomputation cost is $O(\mathcal{T}_{\mathrm{train}} \cdot (e \cdot d + n \cdot d))$ over all training snapshots. In practice, both are a small fraction of the base encoder cost, as the encoder itself involves multi-layer message passing. We report wall-clock training times in Appendix~\ref{app:further}.8.

\section{Additional Experimental Setup} \label{app:implementation}
\subsection{Baselines}
We evaluate nine baseline architectures, grouped into static and dynamic models. For each dynamic backbone (and static backbone with sequential training), we compare the baseline classifier against the same encoder augmented with \proposed.

\noindent{\textbf{Static baselines.}}
\begin{itemize}[leftmargin = 2mm]
    \item \textbf{GCN~\cite{gcn}.} A graph convolutional network. We evaluate two training protocols: (i) \textit{collapsed}, where all training snapshots are merged into a single graph by taking the union of edges across snapshots and the node features (and labels) in the last training snapshot, and (ii) \textit{sequential}, where the model is trained on each training snapshot in order, processing them one at a time. The sequential variant is additionally augmented with \proposed.
    \item \textbf{EERM~\cite{eerm}.} A GNN method designed to handle distribution shift through environment-based invariant learning. Evaluated with the same two training protocols and \proposed~augmentation as GCN.
\end{itemize}

\noindent{\textbf{Dynamic baselines.}}
\begin{itemize}[leftmargin = 2mm]
  \item \textbf{GCN-GRU.} A GCN model co-trained with a GRU on node embeddings. This is a variant of GCRN~\cite{gcrn} adapted in~\cite{evolvegcn}.
    \item \textbf{EvolveGCN~\cite{evolvegcn}.} Uses an RNN~\cite{gru,lstm} to evolve the GCN weight matrices across snapshots. This allows the model to adapt its graph convolution parameters over time without relying on node-level recurrence.
    \item \textbf{DySAT~\cite{dysat}.} Applies structural GAT attention within each snapshot and transformer-style self-attention across the temporal dimension.
    \item \textbf{GCN-SE~\cite{gcnse}.} Applies GCN independently per snapshot and learns squeeze-and-excitation attention weights to combine snapshot-level representations.
    \item \textbf{IDGNN~\cite{idgnn}.} An implicit-depth GNN that iterates message passing until convergence via a fixed-point formulation, avoiding oversmoothing by design.
    \item \textbf{TADGNN~\cite{tadgnn}.} Augments the input graph with time-stamped edges and applies GCNII-style~\cite{gcnii} deep propagation with initial residual connections.
    \item \textbf{SiGNN~\cite{signn}.} Combines GCN with biologically inspired BLIF spiking neurons in a dual-path temporal activation (sigmoid $\times$ spike), designed to capture both smooth and abrupt temporal dynamics.
\end{itemize}

\subsection{Implementation Details}
All models are trained using the Adam~\cite{adam} optimizer. Training runs for a maximum of 500 epochs with early stopping based on validation F1 (patience = 10 epochs). We temporally split the snapshots into training, validation and test snapshots (See Table~\ref{tab:datasets}). For all backbones we follow the originally suggested hyperparameter values. For \proposed, the change detection head is a 2-layer MLP with hidden dimension 32, and the change-aware classifier is a 2-layer MLP with hidden dimension 128. The self-projection $\mathbf{W}_{\mathrm{self}}$ maps from input feature dimension $d$ to hidden dimension 128. We set $\lambda_{\mathrm{hom}} = \lambda_{\mathrm{change}} = 0.1$ and margin $m=0.5$ across all \proposed-augmented experiments (See Appendix~\ref{app:further}.4 for a sensitivity analysis). All models (initial backbones and \proposed-augmented) use the same classifier architecture to ensure that performance differences are attributable to the \proposed~components rather than classifier capacity. All experiments are executed based on PyTorch 2.2.1 with CUDA 11.8, and training is performed on a single NVIDIA RTX A6000 GPU (48GB) and Intel Xeon Gold 6226R CPU @ 2.90GHz (2 sockets × 16 cores, 64 threads total).

\clearpage
\section{Full Numerical Results} \label{app:results}
Table~\ref{tab:main_std} reports the standard deviations corresponding to the main results in Table~\ref{tab:main}. Beyond improving mean performance, \proposed~also reduces variance across seeds: baseline standard deviations are consistently higher than their +\proposed~counterparts, indicating more stable training.

\begin{table}
\caption{Full results with standard deviations (5 seeds) corresponding to Table~\ref{tab:main}. All values in percentage points.}
\label{tab:main_std}
\centering
\resizebox{\textwidth}{!}{%
\begin{tabular}{l llll llll llll}
\toprule
& \multicolumn{4}{c}{\textbf{DBLP-S}} & \multicolumn{4}{c}{\textbf{GossipCop}} & \multicolumn{4}{c}{\textbf{Music}} \\
\cmidrule(lr){2-5} \cmidrule(lr){6-9} \cmidrule(lr){10-13}
\textbf{Model} & Sh.Acc & Sh.F1 & St.Acc & St.F1 & Sh.Acc & Sh.F1 & St.Acc & St.F1 & Sh.Acc & Sh.F1 & St.Acc & St.F1 \\
\midrule
\textbf{GCN} (collapsed) & 73.0$_{\pm \textbf{0.2}}$ & 76.5$_{\pm \textbf{0.4}}$ & 96.0$_{\pm \textbf{0.0}}$ & 97.1$_{\pm \textbf{0.0}}$ & 48.6$_{\pm \textbf{0.9}}$ & 22.1$_{\pm \textbf{4.9}}$ & 69.7$_{\pm \textbf{1.3}}$ & 40.1$_{\pm \textbf{6.5}}$ & 38.1$_{\pm \textbf{0.7}}$ & 23.7$_{\pm \textbf{3.2}}$ & 64.5$_{\pm \textbf{0.7}}$ & 22.9$_{\pm \textbf{0.3}}$ \\
\textbf{GCN} (sequential) & 71.8$_{\pm \textbf{0.7}}$ & 75.5$_{\pm \textbf{1.0}}$ & 96.1$_{\pm \textbf{0.0}}$ & 97.1$_{\pm \textbf{0.0}}$ & 52.8$_{\pm \textbf{0.1}}$ & 25.0$_{\pm \textbf{1.2}}$ & 75.6$_{\pm \textbf{0.1}}$ & 45.3$_{\pm \textbf{0.7}}$ & 46.3$_{\pm \textbf{2.9}}$ & 29.0$_{\pm \textbf{3.5}}$ & 73.6$_{\pm \textbf{1.7}}$ & 45.0$_{\pm \textbf{3.9}}$ \\
\rowcolor{green!8} \quad +\proposed~& 83.2$_{\pm \textbf{0.6}}$ & 85.9$_{\pm \textbf{0.5}}$ & 96.2$_{\pm \textbf{0.1}}$ & 97.2$_{\pm \textbf{0.1}}$ & 69.6$_{\pm \textbf{0.6}}$ & 60.1$_{\pm \textbf{3.8}}$ & 87.7$_{\pm \textbf{1.2}}$ & 76.9$_{\pm \textbf{2.5}}$ & 80.0$_{\pm \textbf{0.8}}$ & 78.6$_{\pm \textbf{0.9}}$ & 93.8$_{\pm \textbf{0.3}}$ & 89.5$_{\pm \textbf{0.5}}$ \\
\textbf{EERM} (collapsed) & 73.5$_{\pm \textbf{0.4}}$ & 77.4$_{\pm \textbf{0.5}}$ & 96.2$_{\pm \textbf{0.0}}$ & 97.3$_{\pm \textbf{0.0}}$ & 47.5$_{\pm \textbf{2.4}}$ & 20.0$_{\pm \textbf{6.5}}$ & 69.6$_{\pm \textbf{1.4}}$ & 44.5$_{\pm \textbf{8.5}}$ & 39.2$_{\pm \textbf{2.3}}$ & 22.4$_{\pm \textbf{0.2}}$ & 64.7$_{\pm \textbf{0.3}}$ & 25.6$_{\pm \textbf{2.4}}$ \\
\textbf{EERM} (seq) & 73.6$_{\pm \textbf{0.5}}$ & 77.3$_{\pm \textbf{0.7}}$ & 96.3$_{\pm \textbf{0.0}}$ & 97.3$_{\pm \textbf{0.0}}$ & 52.5$_{\pm \textbf{2.7}}$ & 23.4$_{\pm \textbf{9.4}}$ & 76.1$_{\pm \textbf{4.9}}$ & 47.1$_{\pm \textbf{3.0}}$ & 44.0$_{\pm \textbf{3.4}}$ & 24.6$_{\pm \textbf{6.8}}$ & 73.8$_{\pm \textbf{2.5}}$ & 45.4$_{\pm \textbf{8.0}}$ \\
\rowcolor{green!8} \quad +\proposed~& 83.1$_{\pm \textbf{0.9}}$ & 85.6$_{\pm \textbf{0.9}}$ & 96.1$_{\pm \textbf{0.1}}$ & 97.2$_{\pm \textbf{0.1}}$ & 70.0$_{\pm \textbf{1.0}}$ & 61.0$_{\pm \textbf{2.3}}$ & 87.7$_{\pm \textbf{1.0}}$ & 76.9$_{\pm \textbf{2.2}}$ & 80.3$_{\pm \textbf{0.8}}$ & 79.3$_{\pm \textbf{1.1}}$ & 93.9$_{\pm \textbf{0.5}}$ & 89.5$_{\pm \textbf{1.0}}$ \\
\midrule
\textbf{GCN-GRU} & 63.9$_{\pm \textbf{1.0}}$ & 68.4$_{\pm \textbf{1.1}}$ & 97.0$_{\pm \textbf{0.0}}$ & 97.8$_{\pm \textbf{0.0}}$ & 58.2$_{\pm \textbf{4.7}}$ & 34.1$_{\pm \textbf{6.5}}$ & 78.3$_{\pm \textbf{3.5}}$ & 53.1$_{\pm \textbf{12.1}}$ & 50.1$_{\pm \textbf{7.0}}$ & 35.2$_{\pm \textbf{7.6}}$ & 79.5$_{\pm \textbf{9.8}}$ & 55.1$_{\pm \textbf{7.7}}$ \\
\rowcolor{green!8} \quad +\proposed~& 82.3$_{\pm \textbf{0.5}}$ & 85.0$_{\pm \textbf{0.6}}$ & 96.3$_{\pm \textbf{0.2}}$ & 97.2$_{\pm \textbf{0.1}}$ & 70.1$_{\pm \textbf{0.9}}$ & 64.4$_{\pm \textbf{1.5}}$ & 87.7$_{\pm \textbf{0.7}}$ & 77.5$_{\pm \textbf{0.9}}$ & 80.1$_{\pm \textbf{1.0}}$ & 79.2$_{\pm \textbf{1.0}}$ & 93.7$_{\pm \textbf{0.4}}$ & 89.6$_{\pm \textbf{0.6}}$ \\
\textbf{EvolveGCN} & 75.7$_{\pm \textbf{0.4}}$ & 79.2$_{\pm \textbf{0.5}}$ & 96.2$_{\pm \textbf{0.0}}$ & 97.2$_{\pm \textbf{0.0}}$ & 70.0$_{\pm \textbf{0.7}}$ & 60.1$_{\pm \textbf{2.8}}$ & 88.1$_{\pm \textbf{2.7}}$ & 78.0$_{\pm \textbf{4.7}}$ & 59.9$_{\pm \textbf{0.7}}$ & 52.2$_{\pm \textbf{2.4}}$ & 86.9$_{\pm \textbf{0.8}}$ & 76.3$_{\pm \textbf{0.7}}$ \\
\rowcolor{green!8} \quad +\proposed~& 82.4$_{\pm \textbf{0.8}}$ & 85.1$_{\pm \textbf{0.7}}$ & 96.2$_{\pm \textbf{0.1}}$ & 97.2$_{\pm \textbf{0.1}}$ & 72.3$_{\pm \textbf{1.5}}$ & 64.4$_{\pm \textbf{2.6}}$ & 87.4$_{\pm \textbf{1.9}}$ & 76.3$_{\pm \textbf{3.9}}$ & 81.0$_{\pm \textbf{1.3}}$ & 80.2$_{\pm \textbf{1.6}}$ & 93.6$_{\pm \textbf{0.7}}$ & 88.8$_{\pm \textbf{1.3}}$ \\
\textbf{SiGNN} & 71.3$_{\pm \textbf{3.6}}$ & 75.3$_{\pm \textbf{4.2}}$ & 87.4$_{\pm \textbf{2.4}}$ & 90.6$_{\pm \textbf{2.1}}$ & 49.1$_{\pm \textbf{0.1}}$ & 28.4$_{\pm \textbf{0.7}}$ & 71.7$_{\pm \textbf{1.4}}$ & 49.8$_{\pm \textbf{7.5}}$ & 71.3$_{\pm \textbf{1.8}}$ & 68.0$_{\pm \textbf{3.3}}$ & 89.9$_{\pm \textbf{0.8}}$ & 81.1$_{\pm \textbf{2.5}}$ \\
\rowcolor{green!8} \quad +\proposed~& 83.2$_{\pm \textbf{0.5}}$ & 85.8$_{\pm \textbf{0.6}}$ & 95.9$_{\pm \textbf{0.3}}$ & 97.0$_{\pm \textbf{0.2}}$ & 71.3$_{\pm \textbf{1.0}}$ & 65.5$_{\pm \textbf{2.8}}$ & 86.1$_{\pm \textbf{0.8}}$ & 73.9$_{\pm \textbf{1.2}}$ & 79.4$_{\pm \textbf{0.7}}$ & 78.2$_{\pm \textbf{0.6}}$ & 93.8$_{\pm \textbf{0.2}}$ & 89.6$_{\pm \textbf{0.5}}$ \\
\textbf{GCN-SE} & 47.1$_{\pm \textbf{0.9}}$ & 49.9$_{\pm \textbf{2.0}}$ & 97.2$_{\pm \textbf{0.0}}$ & 97.9$_{\pm \textbf{0.0}}$ & 41.0$_{\pm \textbf{0.8}}$ & 20.1$_{\pm \textbf{2.1}}$ & 84.2$_{\pm \textbf{0.9}}$ & 75.2$_{\pm \textbf{4.7}}$ & 36.8$_{\pm \textbf{2.3}}$ & 30.3$_{\pm \textbf{2.8}}$ & 76.9$_{\pm \textbf{5.8}}$ & 71.2$_{\pm \textbf{2.4}}$ \\
\rowcolor{green!8} \quad +\proposed~& 79.5$_{\pm \textbf{1.4}}$ & 82.2$_{\pm \textbf{1.2}}$ & 96.4$_{\pm \textbf{0.2}}$ & 97.3$_{\pm \textbf{0.2}}$ & 69.3$_{\pm \textbf{0.9}}$ & 60.8$_{\pm \textbf{1.4}}$ & 86.8$_{\pm \textbf{1.6}}$ & 75.7$_{\pm \textbf{3.3}}$ & 79.1$_{\pm \textbf{1.0}}$ & 77.3$_{\pm \textbf{1.0}}$ & 94.0$_{\pm \textbf{0.1}}$ & 89.5$_{\pm \textbf{0.4}}$ \\
\textbf{DySAT} & 61.2$_{\pm \textbf{2.1}}$ & 67.1$_{\pm \textbf{5.8}}$ & 90.7$_{\pm \textbf{13.0}}$ & 94.0$_{\pm \textbf{7.5}}$ & 49.8$_{\pm \textbf{0.9}}$ & 29.0$_{\pm \textbf{4.0}}$ & 75.8$_{\pm \textbf{3.9}}$ & 61.2$_{\pm \textbf{7.2}}$ & 39.9$_{\pm \textbf{1.7}}$ & 34.8$_{\pm \textbf{6.5}}$ & 70.2$_{\pm \textbf{5.2}}$ & 47.9$_{\pm \textbf{4.9}}$ \\
\rowcolor{green!8} \quad +\proposed~& 81.2$_{\pm \textbf{0.5}}$ & 84.1$_{\pm \textbf{0.5}}$ & 96.5$_{\pm \textbf{0.0}}$ & 97.4$_{\pm \textbf{0.0}}$ & 69.4$_{\pm \textbf{1.2}}$ & 60.6$_{\pm \textbf{3.7}}$ & 84.7$_{\pm \textbf{2.2}}$ & 70.5$_{\pm \textbf{6.0}}$ & 80.3$_{\pm \textbf{1.1}}$ & 79.3$_{\pm \textbf{1.4}}$ & 93.4$_{\pm \textbf{0.2}}$ & 88.8$_{\pm \textbf{0.2}}$ \\
\textbf{IDGNN} & 73.3$_{\pm \textbf{0.7}}$ & 75.9$_{\pm \textbf{0.7}}$ & 96.0$_{\pm \textbf{0.0}}$ & 97.1$_{\pm \textbf{0.0}}$ & 58.0$_{\pm \textbf{1.6}}$ & 46.3$_{\pm \textbf{3.4}}$ & 85.1$_{\pm \textbf{1.7}}$ & 71.4$_{\pm \textbf{4.0}}$ & 67.4$_{\pm \textbf{1.0}}$ & 66.1$_{\pm \textbf{0.6}}$ & 93.9$_{\pm \textbf{0.4}}$ & 88.9$_{\pm \textbf{1.0}}$ \\
\rowcolor{green!8} \quad +\proposed~& 83.3$_{\pm \textbf{0.4}}$ & 86.0$_{\pm \textbf{0.5}}$ & 96.0$_{\pm \textbf{0.0}}$ & 97.1$_{\pm \textbf{0.0}}$ & 66.9$_{\pm \textbf{1.4}}$ & 58.2$_{\pm \textbf{1.4}}$ & 86.1$_{\pm \textbf{1.3}}$ & 73.9$_{\pm \textbf{2.4}}$ & 77.8$_{\pm \textbf{0.9}}$ & 76.0$_{\pm \textbf{1.1}}$ & 93.7$_{\pm \textbf{0.4}}$ & 89.1$_{\pm \textbf{0.9}}$ \\
\textbf{TADGNN} & 76.6$_{\pm \textbf{4.5}}$ & 79.9$_{\pm \textbf{4.2}}$ & 96.7$_{\pm \textbf{0.3}}$ & 97.6$_{\pm \textbf{0.2}}$ & 65.9$_{\pm \textbf{2.0}}$ & 50.6$_{\pm \textbf{4.5}}$ & 84.1$_{\pm \textbf{3.0}}$ & 67.2$_{\pm \textbf{7.7}}$ & 69.1$_{\pm \textbf{7.3}}$ & 62.1$_{\pm \textbf{8.6}}$ & 88.4$_{\pm \textbf{7.4}}$ & 75.1$_{\pm \textbf{6.2}}$ \\
\rowcolor{green!8} \quad +\proposed~& 83.2$_{\pm \textbf{0.2}}$ & 85.9$_{\pm \textbf{0.2}}$ & 96.0$_{\pm \textbf{0.0}}$ & 97.0$_{\pm \textbf{0.0}}$ & 69.8$_{\pm \textbf{3.4}}$ & 61.4$_{\pm \textbf{4.9}}$ & 84.6$_{\pm \textbf{3.4}}$ & 70.4$_{\pm \textbf{8.7}}$ & 80.2$_{\pm \textbf{1.0}}$ & 79.0$_{\pm \textbf{1.1}}$ & 93.6$_{\pm \textbf{0.4}}$ & 89.2$_{\pm \textbf{0.9}}$ \\
\bottomrule
\end{tabular}
}
\end{table}

\section{Further Analysis} \label{app:further}
\begin{table}[t]
\centering
\begin{minipage}[t]{0.48\textwidth}
\centering
\caption{AUROC (\%) of $p_{\mathrm{change}}$ for shifter detection on test snapshots.}
\label{tab:pchange_auroc}
\begin{tabular}{l ccc}
\toprule
\textbf{Backbone} & \textbf{DBLP-S} & \textbf{GossipCop} & \textbf{Music} \\
\midrule
GCN-GRU & 93.1 & 97.0 & 79.5 \\
EvolveGCN & 87.2 & 97.6 & 77.9 \\
SiGNN & 83.7 & 96.9 & 79.1 \\
GCN-SE & 87.3 & 96.9 & 78.8 \\
DySAT & 90.4 & 97.0 & 79.2 \\
IDGNN & 85.6 & 97.4 & 79.1 \\
TADGNN & 84.3 & 97.1 & 81.1 \\
\bottomrule
\end{tabular}
\end{minipage}
\hfill
\begin{minipage}[t]{0.48\textwidth}
\centering
\caption{Shifter F1: \proposed~ vs.\ rule-based variant, averaged across backbones.}
\label{tab:rulebased}
\begin{tabular}{l ccc}
\toprule
\textbf{Method} & \textbf{DBLP-S} & \textbf{GossipCop} & \textbf{Music} \\
\midrule
Baseline & 70.8 & 30.1 & 46.8 \\
Rule-based & 74.7 & 51.4 & 18.6 \\
\textbf{+\proposed}~& \textbf{84.9} & \textbf{62.2} & \textbf{78.4} \\
\bottomrule
\end{tabular}
\end{minipage}
\end{table}

\subsection{Analysis on $p_{\mathrm{change}}$}
\noindent{\textbf{The change detection head's performance.}}
The change detection head is trained to distinguish shifters from stable nodes. Table~\ref{tab:pchange_auroc} reports the AUROC of $p_{\mathrm{change}}$ for shifter detection on test snapshots across all backbones and datasets. Detection performance is consistently high (up to 97.4\%), confirming that the structural signals we analyzed in Section~\ref{analysis} provide a reliable basis for identifying label-shifting nodes.

\noindent{\textbf{\proposed~vs.\ rule-based variant.}}
To further highlight the performance of our change-aware classifier, we additionally compare \proposed~against a rule-based variant: given $p_{\mathrm{change}}$ and the previous prediction, when $p_{\mathrm{change}}>0.5$, this variant simply flips from the old label to the opposite class (binary datasets) or replaces it with a random non-old-label class (multiclass datasets). On the binary datasets \textbf{DBLP-S} and \textbf{GossipCop}, the rule-based variant also outperforms the unmodified baselines (See Table~\ref{tab:rulebased}), further confirming that $p_{\mathrm{change}}$ is a meaningful signal for identifying shifters. However, on the multiclass \textbf{Music} dataset, it degrades considerably: knowing \textit{when} a node has shifted is insufficient without knowing \textit{what} it has shifted to. \proposed~outperforms the rule-based variant in all cases, demonstrating that the change-aware classifier learns not only when to deviate from the previous prediction but how, by integrating $p_{\mathrm{change}}$ with other relevant signals, e.g., $\mathbf{z}_{\mathrm{self},v}^{(t)}$.

\subsection{Improvement by Number of Shifts}
\begin{figure}[t]
\centering
\includegraphics[width=\textwidth]{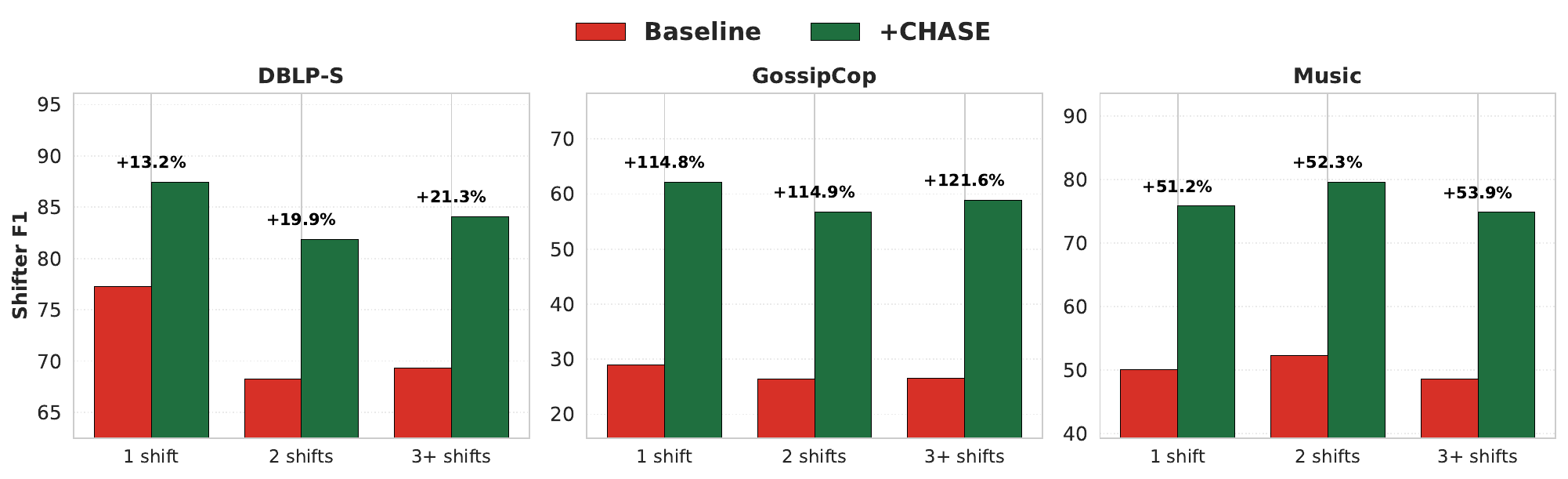}
\vspace{-2ex}
\caption{Relative improvement in shifter F1 (+\proposed~over baseline), stratified by number of label shifts during test snapshots. Averaged across all seven dynamic backbones. \proposed's benefit increases with shift frequency across all datasets.}
\label{fig:num_shifts}
\end{figure}

We stratify nodes by the number of label shifts they undergo during the test window (1, 2, or 3+) and measure \proposed's relative improvement on shifter F1, averaged across all seven dynamic backbones. Fig.~\ref{fig:num_shifts} shows that baseline shifter F1 generally decreases with the number of shifts, confirming that frequently shifting nodes are harder to classify. \proposed's improvement increases accordingly: nodes that shift more frequently benefit the most from \proposed~augmentation.

\begin{figure}[t]
\centering
\includegraphics[width=0.95\textwidth]{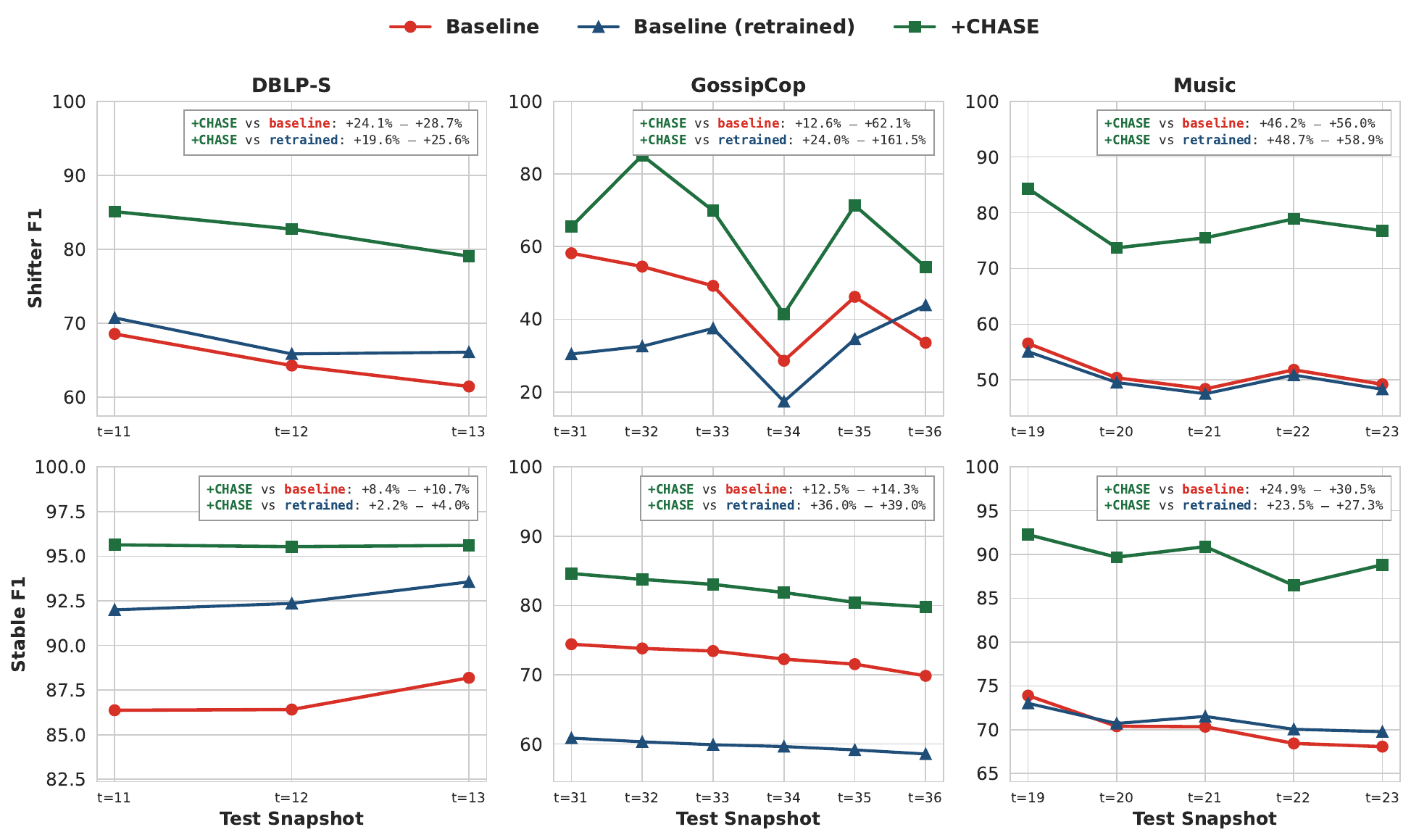}
\vspace{-1ex}
\caption{Shifter and stable F1 across test snapshots for baseline, baseline with retraining, and baseline + \proposed. \proposed~outperforms both alternatives consistently, despite the retraining baseline having access to ground-truth labels from previous test snapshots. Averaged across all backbones.}
\label{fig:consistency}
\vspace{-2ex}
\end{figure}

\subsection{Per-Snapshot Consistency and Comparison with Retraining Baselines}
We additionally compare \proposed~against retraining baselines that re-train the backbone before each new test snapshot using all previously observed data (i.e., up to and including the test snapshot immediately before it), including ground-truth labels. This represents a strictly stronger baseline, as \proposed's structural signals rely only on its own predictions from the previous snapshot. Fig.~\ref{fig:consistency} shows that \proposed~consistently achieves the highest shifter and stable F1 across test snapshots on all datasets, with improvement distributed evenly across time rather than concentrated in particular snapshots. Notably, retraining does not always benefit baselines, and even when it does, \proposed~still outperforms substantially.

\begin{figure}[t]
  \centering
  \includegraphics[width=0.9\textwidth]{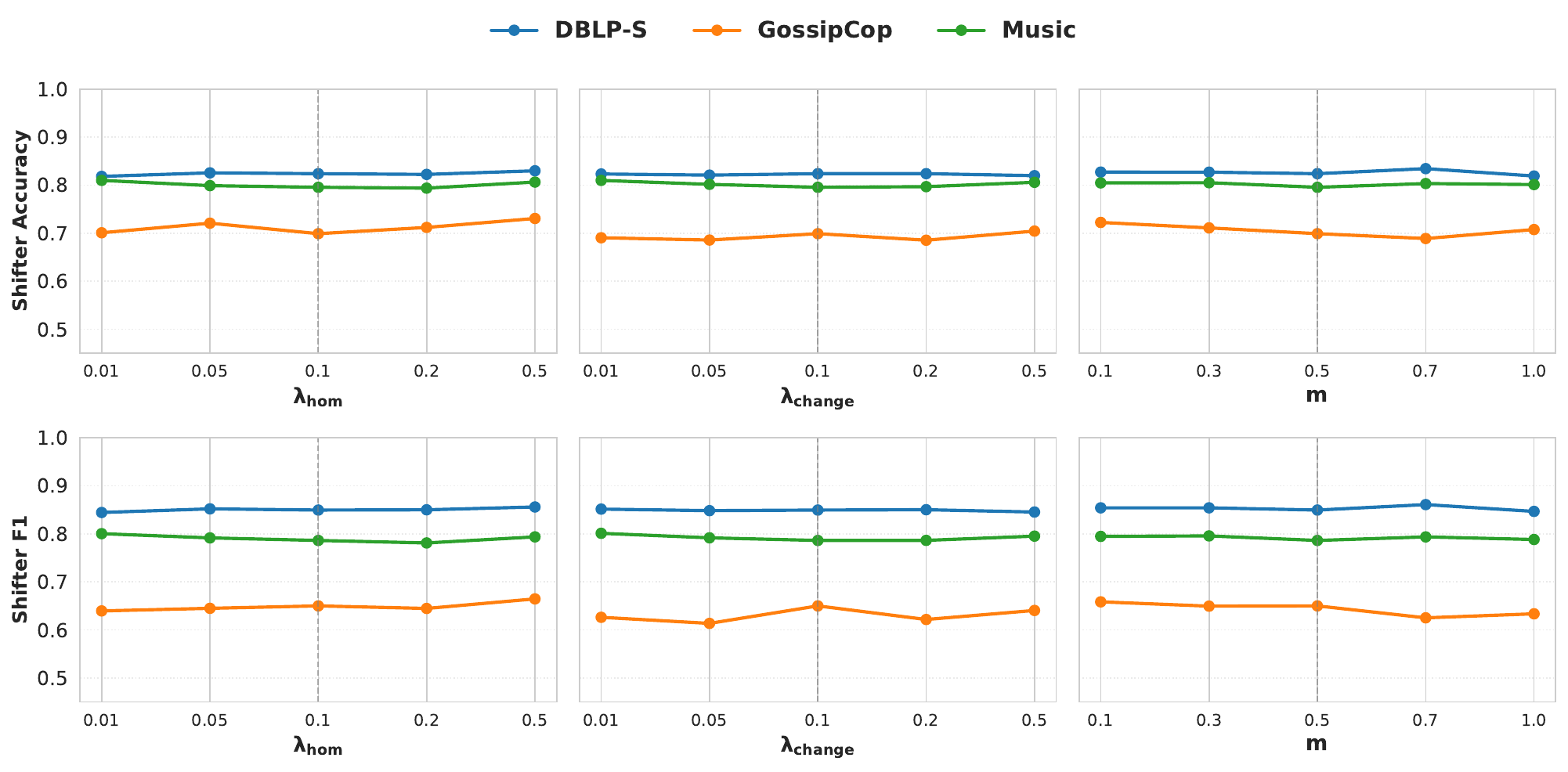}
  \caption{Hyperparameter sensitivity on GCN-GRU. Shifter accuracy and F1 remains stable across all tested values of $\lambda_{\mathrm{hom}}$, $\lambda_{\mathrm{change}}$, and $m$. Dashed lines indicate default values.}
  \label{fig:sensitivity}
\end{figure}

\subsection{Hyperparameter Sensitivity}
\proposed~introduces three hyperparameters: the contrastive loss weight $\lambda_{\mathrm{hom}}$, the change detection loss weight $\lambda_{\mathrm{change}}$, and the contrastive margin $m$ (defaults: $0.1$, $0.1$, $0.5$). We vary each independently while holding the others at their defaults, using GCN-GRU as the backbone across all three datasets. Fig.~\ref{fig:sensitivity} shows that performance is stable across all tested ranges ($\lambda_{\mathrm{hom}}, \lambda_{\mathrm{change}} \in \{0.01, 0.05, 0.1, 0.2, 0.5\}$; $m \in \{0.1, 0.3, 0.5, 0.7, 1.0\}$), indicating that \proposed~is minimally sensitive to hyperparameter tuning.

\begin{table}[t]
\centering
\caption{Comparison of a shifter vs.\ a stable node.}
\label{tab:case}
\begin{tabular}{lcc}
\toprule
&shifter (node 46)&stable (node 688)\\
&\small{label: $0 \rightarrow 1$}&\small{label: $0 \rightarrow 0$}\\
\midrule
$\mathrm{self\_drift}$&91.65&6.78\\
$\mathrm{nbr\_drift}$&121.67&5.70\\
$h_{\mathrm{prev}}$&0.76&0.77\\
$h_{\mathrm{curr}}$&0.05&0.80\\
$p_{\mathrm{change}}$&0.83&0.04\\
\bottomrule
\end{tabular}
\end{table}

\subsection{Case Study}
We trace two nodes through \proposed's signals on Music using GCN-GRU as the backbone. Table~\ref{tab:case} demonstrates that both nodes instantiate the patterns in Fig.~\ref{fig:signals}. Node 46 shows high drift on both signals together with low current and high previous homophily, exactly the condition under which shifters are misclassified, and its embedding is duly anchored to the old class (negative $\Delta_c$), so the baseline predicts the old label. Based on these signals, \proposed's change detection head fires ($p_{\mathrm{change}} = 0.83$), and the change-aware classifier moves the node across to the new class ($\Delta_c$ rises by +0.24, turning positive), predicting correctly. Node 688 matches the stable-node profile, triggers no change signal, and is left correct and untouched.

\subsection{Robustness to Incorrect Previous Predictions}
At test time, \proposed~uses $\hat{\mathbf{y}}^{(t{-}1)}$ from the model's own predictions, which may be incorrect. We assess the impact by grouping shifters into those with correct vs.\ incorrect previous predictions and measuring the change in shifter F1 relative to the baseline, averaged across all backbones (See Table~\ref{tab:wrongpred}).

When $\hat{\mathbf{y}}^{(t{-}1)}$ is correct, \proposed~yields substantial improvement across all datasets, as expected. When $\hat{\mathbf{y}}^{(t{-}1)}$ is incorrect, the effect depends on the label space. On the binary datasets \textbf{DBLP-S} and \textbf{GossipCop}, performance degrades modestly: an incorrect previous prediction causes $h_{\mathrm{proxy}}$ to be computed against the wrong class for the target node, lowering its value and in turn reducing $p_{\mathrm{change}}$, as lower $h_{\mathrm{proxy}}$ signals a less likely change. The model therefore shifts less aggressively from the (wrong) previous prediction, which for binary classification often results in a correct current prediction. On the multiclass \textbf{Music} dataset, \proposed~still yields substantial improvement: with more than two classes, $h_{\mathrm{proxy}}$ is less affected by an incorrect previous prediction, so $p_{\mathrm{change}}$ remains comparable to the correct-prediction group. The classifier is still signaled to shift, and the additional inputs provide sufficient information to identify the correct new class.

\begin{table}
\caption{Effect of incorrect $\hat{\mathbf{y}}^{(t{-}1)}$ on shifters: change in F1 ($\Delta$F1, in percentage points) from baseline, mean $h_{\mathrm{proxy}}$, and mean $p_{\mathrm{change}}$. Averaged across backbones.}
\label{tab:wrongpred}
\centering
\begin{tabular}{l ccc ccc ccc}
\toprule
& \multicolumn{3}{c}{\textbf{DBLP-S}} & \multicolumn{3}{c}{\textbf{GossipCop}} & \multicolumn{3}{c}{\textbf{Music}} \\
\cmidrule(lr){2-4} \cmidrule(lr){5-7} \cmidrule(lr){8-10}
$\hat{\mathbf{y}}^{(t{-}1)}$ & $\Delta$F1 & $h_{\mathrm{proxy}}$ & $p_{\mathrm{change}}$ & $\Delta$F1 & $h_{\mathrm{proxy}}$ & $p_{\mathrm{change}}$ & $\Delta$F1 & $h_{\mathrm{proxy}}$ & $p_{\mathrm{change}}$ \\
\midrule
Correct & +27.7 & 0.707 & 0.750 & +40.0 & 0.910 & 0.909 & +30.9 & 0.442 & 0.664 \\
Incorrect & -4.6 & 0.589 & 0.562 & -8.0 & 0.711 & 0.696 & +22.5 & 0.472 & 0.643 \\
\bottomrule
\end{tabular}
\end{table}

\subsection{Robustness to Input Noise}
We evaluate \proposed's robustness under two types of input perturbation using GCN-GRU as the backbone. \textit{Feature noise} adds Gaussian noise scaled to each feature dimension: $\mathbf{x}^{(t)} \leftarrow \mathbf{x}^{(t)} + \epsilon \cdot \mathrm{std}(\mathbf{x}^{(t)}) \cdot \alpha$, where $\epsilon \sim \mathcal{N}(0, \mathbf{I})$ and $\alpha$ is the noise level. \textit{Edge noise} randomly drops a fraction $\alpha$ of edges per snapshot and adds an equal number of random edges, preserving the total edge count while shuffling neighborhood structure. We test noise levels $\alpha \in \{0.0, 0.1, 0.3, 0.5\}$.

\begin{figure}[t]
  \centering
  \includegraphics[width=\textwidth]{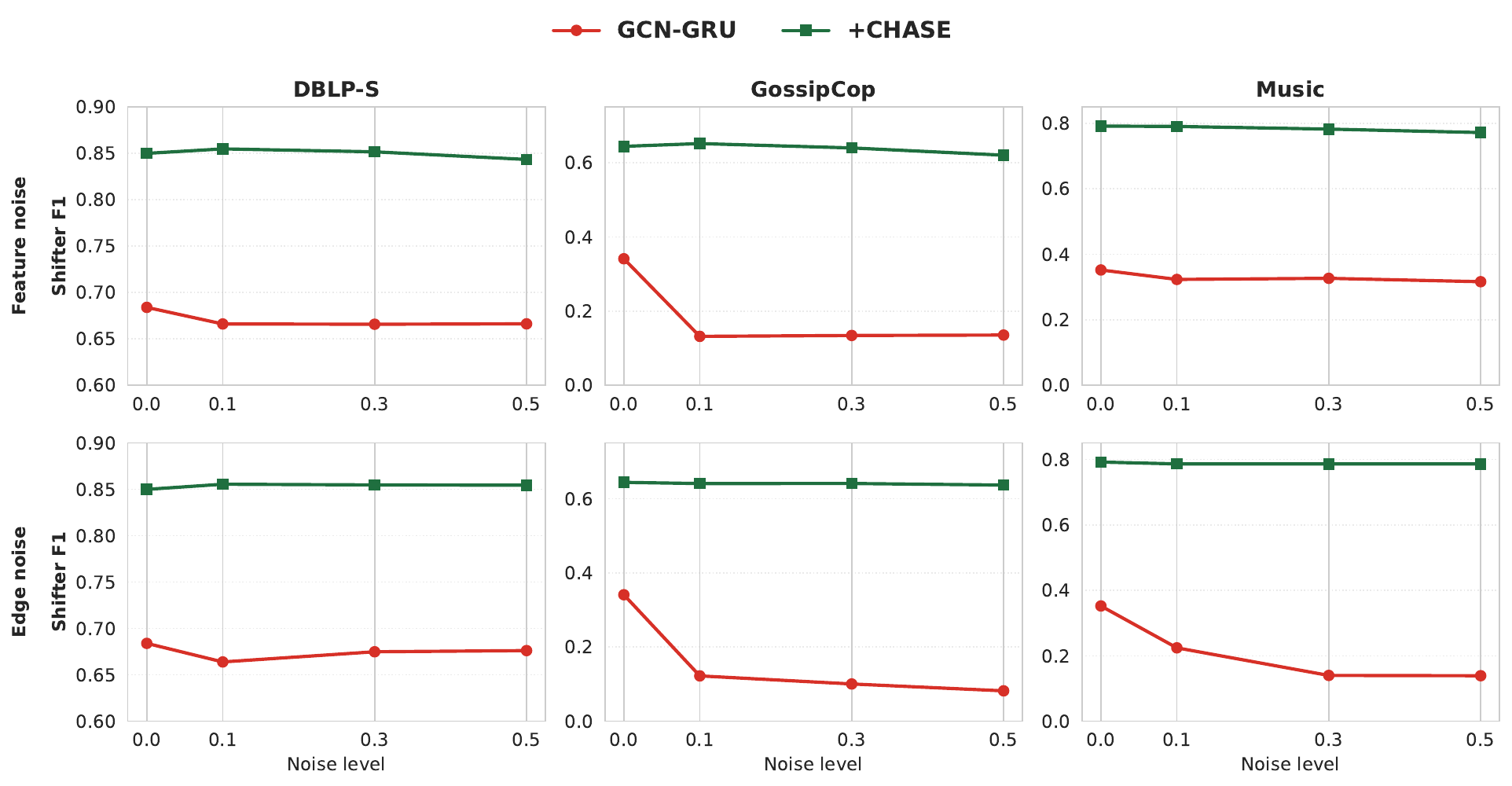}
  \caption{Shifter F1 under increasing feature noise (Top) and edge noise (Bottom) on GCN-GRU. \proposed~'s degradation is minimal compared to the baseline across all noise levels and datasets.}
  \label{fig:noise}
\end{figure}

Fig.~\ref{fig:noise} shows that \proposed~is more robust than the baseline under both perturbation types across all datasets. Even at $\alpha = 0.5$, \proposed~retains a substantial advantage: the structural drift signals (feature drift, neighborhood drift) remain meaningfully discriminative between shifters and stable nodes even under heavy noise (see Table~\ref{tab:noise_drift}), as they measure \textit{relative change} between snapshots rather than absolute values. This robustness to noisy inputs is encouraging for real-world deployment, where node features and graph structure are often incomplete or imprecise.

\begin{table}
\caption{Mean drift signal values for shifters (Sh.) and stable nodes (St.) at noise level $\alpha = 0.5$. Separation persists under both perturbation types.}
\label{tab:noise_drift}
\centering
\resizebox{\textwidth}{!}{%
\begin{tabular}{l cc cc cc cc cc cc}
\toprule
& \multicolumn{4}{c}{\textbf{DBLP-S}} & \multicolumn{4}{c}{\textbf{GossipCop}} & \multicolumn{4}{c}{\textbf{Music}} \\
\cmidrule(lr){2-5} \cmidrule(lr){6-9} \cmidrule(lr){10-13}
& \multicolumn{2}{c}{$\mathrm{self\_drift}$} & \multicolumn{2}{c}{$\mathrm{nbr\_drift}$} & \multicolumn{2}{c}{$\mathrm{self\_drift}$} & \multicolumn{2}{c}{$\mathrm{nbr\_drift}$} & \multicolumn{2}{c}{$\mathrm{self\_drift}$} & \multicolumn{2}{c}{$\mathrm{nbr\_drift}$} \\
\cmidrule(lr){2-3} \cmidrule(lr){4-5} \cmidrule(lr){6-7} \cmidrule(lr){8-9} \cmidrule(lr){10-11} \cmidrule(lr){12-13}
\textbf{Noise} & Sh. & St. & Sh. & St. & Sh. & St. & Sh. & St. & Sh. & St. & Sh. & St. \\
\midrule
Feature & 10.61 & 8.47 & 3.86 & 3.22 & 9.42 & 4.85 & 1.35 & 0.16 & 178.05 & 149.63 & 42.32 & 21.47 \\
Edge & 9.25 & 5.97 & 4.39 & 4.01 & 8.06 & 0.57 & 3.33 & 2.49 & 118.19 & 66.68 & 38.89 & 31.46 \\
\bottomrule
\end{tabular}
}
\end{table}

\begin{table}[t]
\caption{\proposed~overhead: additional wall-clock time (seconds per epoch) and peak GPU memory (MB) over the baseline.}
\label{tab:timing}
\centering
\resizebox{0.9\textwidth}{!}{%
\begin{tabular}{l cc cc cc}
\toprule
& \multicolumn{2}{c}{\textbf{DBLP-S}} & \multicolumn{2}{c}{\textbf{GossipCop}} & \multicolumn{2}{c}{\textbf{Music}} \\
\cmidrule(lr){2-3} \cmidrule(lr){4-5} \cmidrule(lr){6-7}
\textbf{Backbone} & $\Delta$s/ep & $\Delta$MB & $\Delta$s/ep & $\Delta$MB & $\Delta$s/ep & $\Delta$MB \\
\midrule
GCN-GRU & +0.24 & +25.6 & +0.35 & +6.8 & +0.24 & +2.2 \\
EvolveGCN & +0.40 & +42.5 & +0.50 & +12.7 & +0.22 & +7.4 \\
SiGNN & +0.25 & +137.9 & +0.37 & +10.5 & +0.24 & +3.1 \\
GCN-SE & +0.30 & +42.9 & +0.38 & +18.4 & +0.21 & +3.1 \\
DySAT & +0.24 & +184.9 & +0.43 & +85.2 & +0.19 & +8.8 \\
IDGNN & +0.14 & +10.6 & +0.29 & +3.4 & +0.23 & +0.5 \\
TADGNN & +0.33 & +185.8 & +0.36 & +88.5 & +0.23 & +10.5 \\
\bottomrule
\end{tabular}
}
\end{table}

\subsection{Computation Overhead}
\proposed~adds a change detection head, a self-projection layer, and a change-aware classifier MLP on top of the backbone. We benchmark the resulting overhead on a single GPU (5 warmup + 20 measured epochs per configuration). Table~\ref{tab:timing} reports the per-epoch wall-clock and peak GPU memory overhead.

\noindent{\textbf{Per-epoch wall-clock.}}
The absolute overhead is bounded at $\sim$0.24--0.50s per epoch across all configurations. Even over 100 training epochs, this translates to under a minute of additional wall-clock in the worst case.

\noindent{\textbf{Precomputation.}}
Structural signals and homophily weights are computed once per training run over the training and validation snapshots: a one-time cost of 15.1s on DBLP-S (11 snapshots), 8.1s on GossipCop (31 snapshots), and 1.5s on Music (19 snapshots).

\noindent{\textbf{Scalability on massive graphs.}}
To address whether \proposed~scales beyond the three real-world datasets used in the main experiments (35K nodes maximum), we benchmark on a massive synthetic dynamic graph with 1,000,000 nodes, 10 snapshots (split 6/2/2), 2 classes, and average degree $\approx$ 20 per snapshot. Each snapshot's edges are independently sampled from a stochastic block model with 80\% within-class assortativity. Node features are 768-dimensional class-conditional Gaussians, matching our initial datasets. On the GCN-GRU backbone, \proposed~adds 15.2s per epoch (i.e., $\sim$25 min overhead at 100 epochs) and 635MB to peak GPU memory, while the one-time precomputation cost is 377.3s (8 snapshots). These results show that \proposed's overhead is modest relative to standard large-scale dynamic GNN training~\cite{scalability1,scalability2}, where per-epoch training times range from minutes to hours, highlighting its practicality.


\clearpage
\input{checklist.tex}

\end{document}

%% file: checklist.tex
\section*{NeurIPS Paper Checklist}

\begin{enumerate}

\item {\bf Claims}
    \item[] Question: Do the main claims made in the abstract and introduction accurately reflect the paper's contributions and scope?
    \item[] Answer: \answerYes{} 
    \item[] Justification: Yes, the abstract and introduction accurately reflect the paper's contributions and scope.

\item {\bf Limitations}
    \item[] Question: Does the paper discuss the limitations of the work performed by the authors?
    \item[] Answer: \answerYes{} 
    \item[] Justification: Yes, we discuss limitations of our work in the appendix.

\item {\bf Theory assumptions and proofs}
    \item[] Question: For each theoretical result, does the paper provide the full set of assumptions and a complete (and correct) proof?
    \item[] Answer: \answerNA{} 
    \item[] Justification: The paper does not utilize theoretical results.

    \item {\bf Experimental result reproducibility}
    \item[] Question: Does the paper fully disclose all the information needed to reproduce the main experimental results of the paper to the extent that it affects the main claims and/or conclusions of the paper (regardless of whether the code and data are provided or not)?
    \item[] Answer: \answerYes{} 
    \item[] Justification: Yes, we discuss implementation details and pseudocode in the appendix.

\item {\bf Open access to data and code}
    \item[] Question: Does the paper provide open access to the data and code, with sufficient instructions to faithfully reproduce the main experimental results, as described in supplemental material?
    \item[] Answer: \answerYes{} 
    \item[] Justification: Yes, we provide access to the code.

\item {\bf Experimental setting/details}
    \item[] Question: Does the paper specify all the training and test details (e.g., data splits, hyperparameters, how they were chosen, type of optimizer) necessary to understand the results?
    \item[] Answer: \answerYes{} 
    \item[] Justification: Yes, we specify training and test details in the appendix.

\item {\bf Experiment statistical significance}
    \item[] Question: Does the paper report error bars suitably and correctly defined or other appropriate information about the statistical significance of the experiments?
    \item[] Answer: \answerYes{} 
    \item[] Justification: Yes, we provide standard deviations for the main tables.

\item {\bf Experiments compute resources}
    \item[] Question: For each experiment, does the paper provide sufficient information on the computer resources (type of compute workers, memory, time of execution) needed to reproduce the experiments?
    \item[] Answer: \answerYes{} 
    \item[] Justification: Yes, we discuss such details in the appendix.
    
\item {\bf Code of ethics}
    \item[] Question: Does the research conducted in the paper conform, in every respect, with the NeurIPS Code of Ethics \url{https://neurips.cc/public/EthicsGuidelines}?
    \item[] Answer: \answerYes{} 
    \item[] Justification: Yes, we follow the NeurIPS Code of Ethics.

\item {\bf Broader impacts}
    \item[] Question: Does the paper discuss both potential positive societal impacts and negative societal impacts of the work performed?
    \item[] Answer: \answerNA{} 
    \item[] Justification: We do not believe there are potential negative societal impacts of the work performed.
    
\item {\bf Safeguards}
    \item[] Question: Does the paper describe safeguards that have been put in place for responsible release of data or models that have a high risk for misuse (e.g., pre-trained language models, image generators, or scraped datasets)?
    \item[] Answer: \answerNA{} 
    \item[] Justification: This paper poses no such risks.

\item {\bf Licenses for existing assets}
    \item[] Question: Are the creators or original owners of assets (e.g., code, data, models), used in the paper, properly credited and are the license and terms of use explicitly mentioned and properly respected?
    \item[] Answer: \answerYes{} 
    \item[] Justification: Yes, we have adequately cited the relevant works and data sources.

\item {\bf New assets}
    \item[] Question: Are new assets introduced in the paper well documented and is the documentation provided alongside the assets?
    \item[] Answer: \answerYes{} 
    \item[] Justification: Yes, our source code is provided and well documented, and we discuss implementation details alongside dataset construction details in the appendix.

\item {\bf Crowdsourcing and research with human subjects}
    \item[] Question: For crowdsourcing experiments and research with human subjects, does the paper include the full text of instructions given to participants and screenshots, if applicable, as well as details about compensation (if any)? 
    \item[] Answer: \answerNA{} 
    \item[] Justification: The paper does not involve crowdsourcing nor research with human subjects.

\item {\bf Institutional review board (IRB) approvals or equivalent for research with human subjects}
    \item[] Question: Does the paper describe potential risks incurred by study participants, whether such risks were disclosed to the subjects, and whether Institutional Review Board (IRB) approvals (or an equivalent approval/review based on the requirements of your country or institution) were obtained?
    \item[] Answer: \answerNA{} 
    \item[] Justification: The paper does not involve crowdsourcing nor research with human subjects.

\item {\bf Declaration of LLM usage}
    \item[] Question: Does the paper describe the usage of LLMs if it is an important, original, or non-standard component of the core methods in this research? Note that if the LLM is used only for writing, editing, or formatting purposes and does \emph{not} impact the core methodology, scientific rigor, or originality of the research, declaration is not required.
    \item[] Answer: \answerNA{} 
    \item[] Justification: Our model does not involve LLMs.

\end{enumerate}

%% file: reference.bib
@article{entitychange,
  title={Temporal networks},
  author={Holme, Petter and Saram{\"a}ki, Jari},
  journal={Physics reports},
  volume={519},
  number={3},
  pages={97--125},
  year={2012},
  publisher={Elsevier}
}

@article{fakenewsnet,
  title={Fakenewsnet: A data repository with news content, social context, and spatiotemporal information for studying fake news on social media},
  author={Shu, Kai and Mahudeswaran, Deepak and Wang, Suhang and Lee, Dongwon and Liu, Huan},
  journal={Big data},
  volume={8},
  number={3},
  pages={171--188},
  year={2020},
  publisher={SAGE Publications Sage CA: Los Angeles, CA}
}

@article{usernewsexposure,
  title={Exposure to ideologically diverse news and opinion on Facebook},
  author={Bakshy, Eytan and Messing, Solomon and Adamic, Lada A},
  journal={Science},
  volume={348},
  number={6239},
  pages={1130--1132},
  year={2015},
  publisher={American Association for the Advancement of Science}
}

@inproceedings{decor,
  title={Decor: Degree-corrected social graph refinement for fake news detection},
  author={Wu, Jiaying and Hooi, Bryan},
  booktitle={Proceedings of the 29th ACM SIGKDD conference on knowledge discovery and data mining},
  pages={2582--2593},
  year={2023}
}

@article{dgsurvey1,
  title={Representation learning for dynamic graphs: A survey},
  author={Kazemi, Seyed Mehran and Goel, Rishab and Jain, Kshitij and Kobyzev, Ivan and Sethi, Akshay and Forsyth, Peter and Poupart, Pascal},
  journal={Journal of Machine Learning Research},
  volume={21},
  number={70},
  pages={1--73},
  year={2020}
}

@article{dgsurvey2,
  title={A survey of dynamic graph neural networks},
  author={Zheng, Yanping and Yi, Lu and Wei, Zhewei},
  journal={Frontiers of Computer Science},
  volume={19},
  number={6},
  pages={196323},
  year={2025},
  publisher={Springer}
}

@article{gcn,
  title={Semi-supervised classification with graph convolutional networks},
  author={Kipf, Thomas N and Welling, Max},
  journal={arXiv preprint arXiv:1609.02907},
  year={2016}
}

@inproceedings{gru,
  title={Learning phrase representations using RNN encoder--decoder for statistical machine translation},
  author={Cho, Kyunghyun and Van Merri{\"e}nboer, Bart and Gul{\c{c}}ehre, {\c{C}}a{\u{g}}lar and Bahdanau, Dzmitry and Bougares, Fethi and Schwenk, Holger and Bengio, Yoshua},
  booktitle={Proceedings of the 2014 conference on empirical methods in natural language processing (EMNLP)},
  pages={1724--1734},
  year={2014}
}

@article{lstm,
  title={Long short-term memory},
  author={Hochreiter, Sepp and Schmidhuber, J{\"u}rgen},
  journal={Neural computation},
  volume={9},
  number={8},
  pages={1735--1780},
  year={1997},
  publisher={MIT press}
}

@article{gat,
  title={Graph attention networks},
  author={Veli{\v{c}}kovi{\'c}, Petar and Cucurull, Guillem and Casanova, Arantxa and Romero, Adriana and Lio, Pietro and Bengio, Yoshua},
  journal={arXiv preprint arXiv:1710.10903},
  year={2017}
}

@inproceedings{gcrn,
  title={Structured sequence modeling with graph convolutional recurrent networks},
  author={Seo, Youngjoo and Defferrard, Micha{\"e}l and Vandergheynst, Pierre and Bresson, Xavier},
  booktitle={International conference on neural information processing},
  pages={362--373},
  year={2018},
  organization={Springer}
}

@inproceedings{evolvegcn,
  title={Evolvegcn: Evolving graph convolutional networks for dynamic graphs},
  author={Pareja, Aldo and Domeniconi, Giacomo and Chen, Jie and Ma, Tengfei and Suzumura, Toyotaro and Kanezashi, Hiroki and Kaler, Tim and Schardl, Tao and Leiserson, Charles},
  booktitle={Proceedings of the AAAI conference on artificial intelligence},
  volume={34},
  number={04},
  pages={5363--5370},
  year={2020}
}

@inproceedings{dysat,
  title={Dysat: Deep neural representation learning on dynamic graphs via self-attention networks},
  author={Sankar, Aravind and Wu, Yanhong and Gou, Liang and Zhang, Wei and Yang, Hao},
  booktitle={Proceedings of the 13th international conference on web search and data mining},
  pages={519--527},
  year={2020}
}

@inproceedings{gcnse,
  title={Gcn-se: Attention as explainability for node classification in dynamic graphs},
  author={Fan, Yucai and Yao, Yuhang and Joe-Wong, Carlee},
  booktitle={2021 IEEE International Conference on Data Mining (ICDM)},
  pages={1060--1065},
  year={2021},
  organization={IEEE}
}

@inproceedings{idgnn,
  title={Efficient and effective implicit dynamic graph neural network},
  author={Zhong, Yongjian and Vu, Hieu and Yang, Tianbao and Adhikari, Bijaya},
  booktitle={Proceedings of the 30th acm sigkdd conference on knowledge discovery and data mining},
  pages={4595--4606},
  year={2024}
}

@inproceedings{tadgnn,
  title={Dynamic graph node classification via time augmentation},
  author={Sun, Jiarui and Gu, Mengting and Yeh, Chin-Chia Michael and Fan, Yujie and Chowdhary, Girish and Zhang, Wei},
  booktitle={2022 IEEE International Conference on Big Data (Big Data)},
  pages={800--805},
  year={2022},
  organization={IEEE}
}

@article{signn,
  title={SiGNN: A spike-induced graph neural network for dynamic graph representation learning},
  author={Chen, Dong and Zheng, Shuai and Xu, Muhao and Zhu, Zhenfeng and Zhao, Yao},
  journal={Pattern Recognition},
  volume={158},
  pages={111026},
  year={2025},
  publisher={Elsevier}
}

@inproceedings{hydg,
  title={Hypergraph-based dynamic graph node classification},
  author={Ma, Xiaoxu and Zhao, Chen and Shao, Minglai and Lin, Yujie},
  booktitle={ICASSP 2025-2025 IEEE International Conference on Acoustics, Speech and Signal Processing (ICASSP)},
  pages={1--5},
  year={2025},
  organization={IEEE}
}

@article{tgb,
  title={Temporal graph benchmark for machine learning on temporal graphs},
  author={Huang, Shenyang and Poursafaei, Farimah and Danovitch, Jacob and Fey, Matthias and Hu, Weihua and Rossi, Emanuele and Leskovec, Jure and Bronstein, Michael and Rabusseau, Guillaume and Rabbany, Reihaneh},
  journal={Advances in Neural Information Processing Systems},
  volume={36},
  pages={2056--2073},
  year={2023}
}

@article{disshiftsurvey,
  title={Beyond generalization: A survey of out-of-distribution adaptation on graphs},
  author={Liu, Shuhan and Ding, Kaize},
  journal={arXiv preprint arXiv:2402.11153},
  year={2024}
}

@article{disshift1,
  title={Generalizing graph neural networks on out-of-distribution graphs},
  author={Fan, Shaohua and Wang, Xiao and Shi, Chuan and Cui, Peng and Wang, Bai},
  journal={IEEE transactions on pattern analysis and machine intelligence},
  volume={46},
  number={1},
  pages={322--337},
  year={2023},
  publisher={IEEE}
}

@inproceedings{disshift2,
  title={Graph out-of-distribution generalization via causal intervention},
  author={Wu, Qitian and Nie, Fan and Yang, Chenxiao and Bao, Tianyi and Yan, Junchi},
  booktitle={Proceedings of the ACM Web Conference 2024},
  pages={850--860},
  year={2024}
}

@article{disshift3,
  title={Shift-robust gnns: Overcoming the limitations of localized graph training data},
  author={Zhu, Qi and Ponomareva, Natalia and Han, Jiawei and Perozzi, Bryan},
  journal={Advances in Neural Information Processing Systems},
  volume={34},
  pages={27965--27977},
  year={2021}
}

@article{eerm,
  title={Handling distribution shifts on graphs: An invariance perspective},
  author={Wu, Qitian and Zhang, Hengrui and Yan, Junchi and Wipf, David},
  journal={arXiv preprint arXiv:2202.02466},
  year={2022}
}

@inproceedings{gcnii,
  title={Simple and deep graph convolutional networks},
  author={Chen, Ming and Wei, Zhewei and Huang, Zengfeng and Ding, Bolin and Li, Yaliang},
  booktitle={International conference on machine learning},
  pages={1725--1735},
  year={2020},
  organization={PMLR}
}

@InProceedings{adam,
  author    = {Kingma, Diederik and Ba, Jimmy},
  booktitle = {International Conference on Learning Representations (ICLR)},
  title     = {Adam: A Method for Stochastic Optimization},
  year      = {2015},
  address   = {San Diega, CA, USA},
  optmonth  = {12},
}

@article{bert,
  title={Bert: Pre-training of deep bidirectional transformers for language understanding},
  author={Devlin, Jacob and Chang, Ming-Wei and Lee, Kenton and Toutanova, Kristina},
  journal={arXiv preprint arXiv:1810.04805},
  year={2018}
}

@inproceedings{upfd,
  title={User preference-aware fake news detection},
  author={Dou, Yingtong and Shu, Kai and Xia, Congying and Yu, Philip S and Sun, Lichao},
  booktitle={Proceedings of the 44th international ACM SIGIR conference on research and development in information retrieval},
  pages={2051--2055},
  year={2021}
}

@inproceedings{lastfm1,
  title={Predicting dynamic embedding trajectory in temporal interaction networks},
  author={Kumar, Srijan and Zhang, Xikun and Leskovec, Jure},
  booktitle={Proceedings of the 25th ACM SIGKDD international conference on knowledge discovery \& data mining},
  pages={1269--1278},
  year={2019}
}

@inproceedings{lastfm2,
  title={Fast ALS-based tensor factorization for context-aware recommendation from implicit feedback},
  author={Hidasi, Bal{\'a}zs and Tikk, Domonkos},
  booktitle={Joint European Conference on Machine Learning and Knowledge Discovery in Databases},
  pages={67--82},
  year={2012},
  organization={Springer}
}

@article{millionsong,
  title={The million song dataset},
  author={Bertin-Mahieux, Thierry and Ellis, Daniel PW and Whitman, Brian and Lamere, Paul},
  year={2011}
}

@article{tgn,
  title={Temporal graph networks for deep learning on dynamic graphs},
  author={Rossi, Emanuele and Chamberlain, Ben and Frasca, Fabrizio and Eynard, Davide and Monti, Federico and Bronstein, Michael},
  journal={arXiv preprint arXiv:2006.10637},
  year={2020}
}

@inproceedings{dyrep,
  title={Dyrep: Learning representations over dynamic graphs},
  author={Trivedi, Rakshit and Farajtabar, Mehrdad and Biswal, Prasenjeet and Zha, Hongyuan},
  booktitle={International conference on learning representations},
  year={2019}
}

@article{scalability1,
  title={Tgl: A general framework for temporal gnn training on billion-scale graphs},
  author={Zhou, Hongkuan and Zheng, Da and Nisa, Israt and Ioannidis, Vasileios and Song, Xiang and Karypis, George},
  journal={arXiv preprint arXiv:2203.14883},
  year={2022}
}

@article{scalability2,
  title={Etc: Efficient training of temporal graph neural networks over large-scale dynamic graphs},
  author={Gao, Shihong and Li, Yiming and Shen, Yanyan and Shao, Yingxia and Chen, Lei},
  journal={Proceedings of the VLDB Endowment},
  volume={17},
  number={5},
  pages={1060--1072},
  year={2024},
  publisher={VLDB Endowment}
}
